\PassOptionsToPackage{table}{xcolor}
\documentclass{easychair}
\usepackage[T1]{fontenc}
\usepackage[utf8]{inputenc}
\usepackage{times}
\usepackage{latexsym}
\usepackage{microtype}
\usepackage{amsmath}
\usepackage{array}
\usepackage{booktabs}
\usepackage{multirow}
\usepackage{graphicx}
\usepackage{subcaption}

\AtBeginDocument{
  \setlength{\abovedisplayskip}{4pt plus 1pt minus 1pt}
  \setlength{\belowdisplayskip}{4pt plus 1pt minus 1pt}
  \setlength{\abovedisplayshortskip}{3pt plus 1pt minus 1pt}
  \setlength{\belowdisplayshortskip}{3pt plus 1pt minus 1pt}
}
\usepackage{listings}
\usepackage{url}
\usepackage[numbers]{natbib}
\usepackage{float}
\usepackage{placeins}
\usepackage{tipa}
\usepackage{mdframed}
\usepackage{xcolor}
\usepackage{adjustbox}
\usepackage{longtable}
\usepackage{pgfplots}
\usepackage{tikz}
\pgfplotsset{compat=1.18}
\usepgfplotslibrary{groupplots}

\newcommand{\negval}[1]{\textcolor{red!65!black}{#1}}

\usepackage{pgfplots}
\pgfplotsset{compat=1.18}
\usepackage{amssymb}

\usepackage{newunicodechar}
\newunicodechar{ı}{\i}
\newunicodechar{İ}{\.{I}}
\newunicodechar{ğ}{\u{g}}
\newunicodechar{Ğ}{\u{G}}
\newunicodechar{ş}{\c{s}}
\newunicodechar{Ş}{\c{S}}
\newunicodechar{ç}{\c{c}}
\newunicodechar{Ç}{\c{C}}
\newunicodechar{ö}{\"{o}}
\newunicodechar{Ö}{\"{O}}
\newunicodechar{ü}{\"{u}}
\newunicodechar{Ü}{\"{U}}
\newunicodechar{Ə}{\rotatebox[origin=c]{180}{E}}
\newunicodechar{ə}{\textschwa}
\newunicodechar{’}{'}

\title{Cross-Lingual Transfer for Machine Translation in \newline Turkic Languages}

\author{
\scriptsize
\begin{tabular}{ccc}
\textbf{Omer Burak Cinar} &
\textbf{Mehmet Mert Dalkilic} &
\textbf{Cagri Toraman} \\
Middle East Technical University &
Middle East Technical University &
Middle East Technical University \\
Computer Engineering Department &
Computer Engineering Department &
Computer Engineering Department \\
\texttt{cinar.burak@metu.edu.tr} &
\texttt{mert.dalkilic@metu.edu.tr} &
\texttt{ctoraman@metu.edu.tr}
\end{tabular}
}
\institute{}

\authorrunning{Cinar et al.}

\titlerunning{Cross-Lingual Transfer for MT in Turkic Languages}

\begin{document}
\maketitle

\begin{abstract}
Cross-lingual transfer is central to low-resource machine translation, but its behavior within closely related language families remains insufficiently characterized. We study transfer among five Turkic languages---Turkish, Azerbaijani, Uzbek, Kazakh, and Kyrgyz---using pairwise transfer matrices. In this setting, each model is fine-tuned with one transfer source and evaluated on a different transfer target while the translation target remains the same. Across mT5 experiments, we find that transfer is strongest between closely related Turkic pairs, especially Turkish--Azerbaijani and Kazakh--Kyrgyz. We also show that transfer direction matters, and that the same transfer source--transfer target pair can behave differently when the translation target changes. Latinization improves BLEU and chrF in several script-mismatched settings, but its effect is not uniform across metrics. Additional analyses show that transfer sources are mostly stable across different datasets and model settings.
\end{abstract}

\section{Introduction}

Neural Machine Translation (NMT) \cite{sutskever2014sequence} has largely replaced earlier statistical and rule-based machine translation systems. Attention mechanisms \cite{bahdanau2015neural} and the Transformer architecture \cite{vaswani2017attention} have substantially improved translation quality, but low-resource settings still remain difficult because high-quality parallel data is scarce \cite{koehn-knowles-2017-six}. Transfer learning is therefore widely used to share knowledge across languages \cite{zoph-etal-2016-transfer}, especially when languages are typologically related and share vocabulary or morphology \cite{kocmi2020exploring}.

The Turkic family is an unusually useful test bed for pairwise transfer. It combines typological relatedness with noticeable asymmetries in resource level and script: Turkish and Azerbaijani are written in Latin script, while Kazakh and Kyrgyz are commonly written in Cyrillic; Uzbek is mixed in practice but often processed in Latin. This combination creates a controlled setting in which transfer can be studied across related languages while still exposing the effects of script alignment and resource imbalance.

Much existing evidence in multilingual MT comes from English-centric settings, where English is often used as the main source, target, or pivot language in training and evaluation \cite{fan2021beyond}. When English is used as the fixed target language in evaluation pairs, transfer from one source language to another is observed only under English as the target. This does not show how the choice of target language changes cross-lingual transfer. To study this effect within the Turkic family, we use Turkic languages as target languages and ask a more fine-grained question: for a fixed target language, which source language is the best fine-tuning donor for each unseen evaluation source, and does that preference remain stable when the evaluation source, script representation, dataset, or pretraining regime changes? Although target-aware data selection has been studied in multilingual NMT \cite{wang-neubig-2019-target}, pairwise transfer between source languages under different Turkic target languages remains less explored. We address this gap by constructing fixed-target transfer matrices among five Turkic languages.

Throughout the paper, we distinguish between translation direction and transfer direction. The \textbf{translation source} is the language from which a sentence is translated, and the \textbf{translation target} is the language into which it is translated. For example, in Turkish-to-Azerbaijani translation, Turkish is the translation source and Azerbaijani is the translation target. In our fixed-target transfer setting, we additionally define a \textbf{transfer source} and a \textbf{transfer target}. The transfer source is the source language used during fine-tuning, while the transfer target is the source language used only at evaluation time. For instance, if a model is fine-tuned on Turkish$\rightarrow$Azerbaijani and evaluated on Uzbek$\rightarrow$Azerbaijani, then Turkish is the transfer source, Uzbek is the transfer target, and Azerbaijani remains the fixed translation target. This design measures the contribution of the transfer source to the transfer target under the same translation target.
\\
\\
\noindent Our study is organized around three research questions:

\noindent\textbf{RQ1.} Which Turkic languages transfer most effectively to unseen languages in cross-lingual machine translation?

\noindent\textbf{RQ2.} How does the choice of translation target affect cross-lingual transfer?

\noindent\textbf{RQ3.} How does transliteration into the Latin alphabet affect Turkic cross-lingual transfer?
\\
\\
\noindent Our main contributions are as follows:

\noindent\textbf{C1.} To the best of our knowledge, we present the first MT-based pairwise transfer matrix for five Turkic languages.

\noindent\textbf{C2.} Unlike prior transfer studies that rely on non-Turkic pivot languages, we use Turkic languages as both translation sources and targets, enabling a controlled analysis of transfer behavior within the same language family.

\noindent\textbf{C3.} We release our best-performing Turkic machine translation models together with the book-aligned parallel pairs created for this study.\footnote{https://github.com/TurkicTransfer/Cross-LingualTransferforMachineTranslationinTurkicLanguages}

\section{Related Work}
Research in multilingual neural machine translation (MNMT) has extensively studied how transfer emerges across languages, when it is beneficial or harmful, and how data composition shapes these dynamics. We summarize the literature along three axes: (i) transfer learning and cross-lingual transfer, (ii) multilingual interference and directionality, and (iii) machine translation for low-resource and Turkic languages.

\subsection{Transfer Learning and Cross-Lingual Transfer}
Transfer learning plays a central role in low-resource machine translation. \citet{zoph-etal-2016-transfer} show that transferring parameters from a high-resource parent model to a low-resource child model yields significant improvements. Subsequent work demonstrates that transfer is strengthened by linguistic similarity and shared sub-word vocabularies \citep{nguyen-chiang-2017-transfer}, and that even simple continued training can yield strong gains without architectural changes \citep{kocmi-bojar-2018-trivial}. Later analyses confirm that transfer is most effective for low-resource and closely related language pairs \citep{kocmi2020exploring}.

Recent work has focused on measuring transfer more explicitly. Representational Transfer Potential (RTP) \citep{stap-etal-2023-viewing} captures positive and negative transfer via representation similarity rather than surface metrics. Large-scale studies such as ATLAS \citep{longpre2025atlas} and Interference Matrix \citep{alastruey2025interference} further show that transfer is asymmetric and strongly influenced by language family and script similarity.

\subsection{Multilingual Interference and Directionality}
Although multilingual training enables parameter sharing across languages, adding more languages or mixing imbalanced data does not always improve performance. Prior work has identified capacity limitations and negative transfer in multilingual models, often described as the curse of multilinguality \citep{conneau-etal-2020-unsupervised}. This issue is especially relevant for low-resource language families, where dominant high-resource languages may help some transfer directions while hurting others because of inter-language parameter competition \citep{blevins-etal-2024-breaking, liu-niehues-2025-conditions}.

Directionality is also important for cross-lingual transfer. Prior work on joint multilingual training shows that translation tasks can interact positively, negatively, or asymmetrically, where one task improves while the other degrades \citep{wang-zhang-2022-addressing}. Other studies investigate target-aware data selection in multilingual NMT \citep{wang-neubig-2019-target} and the role of translation direction in multilingual training \citep{luo2025beyond}. However, less attention has been given to the direction from transfer source to transfer target, especially how this direction changes under different translation targets. Our work focuses on this setting by constructing pairwise transfer matrices among Turkic languages.

\subsection{Low-Resource and Turkic Machine Translation}
Low-resource machine translation is often constrained by the scarcity of high-quality parallel data. Backtranslation \citep{sennrich-etal-2016-improving} is commonly used to create synthetic parallel data by translating monolingual target-language data into the source language, and high-resource pivot languages such as English are also used to support data construction \citep{kim-etal-2019-pivot}. Beyond data creation, low-resource MT also depends on methods for finding and assessing useful parallel sentence pairs. 
Sentence embedding models such as LaBSE are effective for low-resource bitext mining \citep{chimoto-bassett-2022-low}, while large language models can support translation quality assessment and filtering decisions \citep{kocmi-federmann-2023-gemba}.

Turkic languages present a particularly challenging low-resource setting because many language pairs have limited direct parallel data, and the family also includes differences in script and resource availability. Prior work has demonstrated the usefulness of Turkish as a high-resource pivot for improving NLU performance across other low-resource Turkic languages \citep{senel-etal-2024-kardes}. In machine translation, multilingual training for Turkic languages has also been shown to improve performance through joint training \citep{mirzakhalov-etal-2021-large}. Building on this line of research, we provide a comprehensive cross-lingual analysis of Turkic MT transfer, examining how transfer patterns change across Latinization, model architectures, evaluation datasets, and continual pretraining settings.

\section{Data}

We use three types of data: monolingual corpora for continual pretraining (CPT), bilingual corpora for fine-tuning, and held-out evaluation datasets. The study covers Turkish (\texttt{tr}), Azerbaijani (\texttt{az}), Uzbek (\texttt{uz}), Kazakh (\texttt{kk}), and Kyrgyz (\texttt{ky}), denoted by their respective ISO 639-1 language codes.

For CPT, we collect monolingual data from publicly available sources, including Wikipedia, CC-100 \citep{conneau-etal-2020-unsupervised}, Leipzig Corpora \citep{goldhahn-etal-2012-building}, OSCAR \citep{abadji-etal-2022-towards}, mC4 \citep{xue-etal-2021-mt5}, MADLAD-400 \citep{kudugunta2023madlad400}, and HPLT \citep{aulamo-etal-2023-hplt}. These corpora are used to adapt \texttt{mT5-small} to Turkic text before translation fine-tuning.

For bilingual fine-tuning, we combine publicly available parallel data with synthetic and mined parallel pairs. Public resources include KazParC \citep{yeshpanov-etal-2024-kazparc}, NTREX \citep{federmann-etal-2022-ntrex}, FLORES+ \citep{gordeev-etal-2024-flores}, and OPUS \citep{tiedemann-2012-parallel}. Because direct Turkic--Turkic parallel data is highly imbalanced and not very suitable for training due to short sentence length, we additionally construct training pairs through back-translation with \texttt{facebook/nllb-200-distilled-600M} \citep{nllb2022} and LaBSE-based bitext mining from multilingual book translations \citep{feng-etal-2022-language}. These data are used for fixed-target fine-tuning and transfer-matrix construction.

This data setting is imperfect but realistic for low-resource Turkic MT. Since many fine-tuning pairs are synthetic, we interpret the results primarily as evidence about \emph{relative transfer structure} under a controlled training pipeline, not as a claim that the absolute MT quality is optimal. We discuss this limitation explicitly in the Limitations section.

For evaluation, we use XWMT as the main benchmark because it provides comparable test pairs across the five Turkic languages \citep{mirzakhalov-etal-2021-large}. We use Tatoeba \citep{tiedemann-2020-tatoeba} as a secondary benchmark to test whether donor preferences and transfer patterns remain stable across datasets. Detailed information about data collection, preprocessing, filtering, Latinization, dataset sizes, back-translation, book alignment, and evaluation-set construction is provided in Appendix~\ref{app:dataset-details}.

\section{Methodology}

\paragraph{Continual Pretraining (CPT)}
We build on \texttt{mT5-small} \cite{xue-etal-2021-mt5}, a 300M-parameter multilingual extension of T5 pretrained with a span-corruption objective on mC4. We continue pretraining on Turkic monolingual corpora with the same objective to adapt the model to Turkic text before translation fine-tuning. This follows prior work showing that continued pretraining can improve low-resource translation performance for Uyghur and language adaptation for Turkish \citep{lu-etal-2025-low,toraman-2024-adapting}. In the allCPT setting, we sample one million approximately 512-token chunks from each of the five languages. Each training sample is prepended with a language tag, such as \texttt{<tr>}, to preserve language identity.

\paragraph{Fine-tuning}
For each fine-tuning run, we select one translation pair $(i \rightarrow t)$, where $i$ is the source language and $t$ is the target language. We use explicit translation-source and translation-target tags, following the text-to-text formulation of T5-style models \citep{Raffel:2020} and multilingual translation tagging conventions \citep{johnson-etal-2017-googles}. Each input is formatted as:
\begin{equation}
\texttt{<src\_lang> <tgt\_lang>: src sentence}
\end{equation}
where \texttt{<src\_lang>} denotes the translation source and \texttt{<tgt\_lang>} denotes the translation target. Both are one of \texttt{<tr>}, \texttt{<az>}, \texttt{<uz>}, \texttt{<kk>}, and \texttt{<ky>}.

\paragraph{Evaluation}
\label{par:evaluation-protocol}
Evaluation is conducted in a zero-shot and fixed-translation-target setting. After fine-tuning on one translation pair $(i \rightarrow t)$, we evaluate the model on translation pairs $(j \rightarrow t)$, where the translation target $t$ remains fixed and $j$ varies over the  Turkic languages. 

When $i \neq j$, $i$ is the \textit{transfer source} and $j$ is the \textit{transfer target}: by keeping the translation target fixed, we measure how fine-tuning on $i \rightarrow t$ affects performance on the unseen pair $j \rightarrow t$. In other words, we measure how much the transfer source $i$ contributes to the transfer target $j$ when the translation target is $t$.

When $i=j$, the model is evaluated on the same translation pair used during fine-tuning. 
This is the easiest evaluation case, because the model has seen fine-tuning data for the exact source--target direction being tested. We use this as a reference score for the transfer cases where $i \neq j$. 

\paragraph{Recovery rate}
To quantify cross-lingual transfer, we use recovery rate adapted from prior transfer-analysis work \citep{turc-etal-2021-revisiting, toraman-etal-2022-large}. For an evaluation metric $m$, let $S_m(i,j,t)$ denote the score of a model fine-tuned on $i \rightarrow t$ and evaluated on $j \rightarrow t$, where $i$ is the transfer source, $j$ is the transfer target, and $t$ is the translation target. Recovery rate measures the percentage of performance retained when the transfer source $i$ is different from the transfer target $j$ for a certain translation target $t$. It is computed as:

\begin{equation}
R_m(i,j,t)=
\frac{S_m(i,j,t)}{S_m(j,j,t)} \times 100 .
\end{equation}

To obtain a language-pair-level transfer score from transfer source $i$ to transfer target $j$, we average recovery over all valid translation targets that are different from both $i$ and $j$:

\begin{equation}
T_m(i,j)=
\frac{1}{|\mathcal{L}\setminus\{i,j\}|}
\sum_{t\in \mathcal{L}\setminus\{i,j\}} R_m(i,j,t),
\end{equation}

where $\mathcal{L}=\{\mathrm{tr},\mathrm{az},\mathrm{uz},\mathrm{kk},\mathrm{ky}\}$. 
Thus, $T_m(i,j)$ summarizes how strongly transfer source $i$ supports transfer target $j$ across possible Turkic translation targets. 

We also include English as a high-resource external transfer source from a different language family. English provides a useful contrast for observing how transfer behaves when the fine-tuning source is not linguistically close to the Turkic languages. For Turkic transfer sources, recovery is averaged over translation targets $t \in \mathcal{L}\setminus\{i,j\}$, so that the translation target differs from both the transfer source $i$ and the transfer target $j$. For English, this condition is relaxed because English is included only as an external transfer source and not as a translation target in $\mathcal{L}$. Therefore, the English row is averaged over translation targets $t \in \mathcal{L}\setminus\{j\}$.

\paragraph{Evaluation Metrics}
We report four evaluation metrics. BLEU \citep{papineni-etal-2002-bleu} measures word-level n-gram overlap between the model output and the reference translation, while chrF \citep{popovic-2015-chrf} measures character-level n-gram overlap with an F-score. We also use neural evaluation metrics: COMET \citep{rei-etal-2020-comet} is a reference-based metric that estimates translation quality using the source sentence, model output, and reference translation, whereas COMETKiwi \citep{rei-etal-2022-cometkiwi} is a reference-free quality estimation metric that evaluates the model output using only the source sentence. We report all metrics consistently across the transfer-matrix experiments.

\section{Experiments and Results}
\label{sec:experiments-results}

\subsection{Cross-Lingual Transfer Matrices (RQ1)}
\label{sec:transfer-matrices}
We first evaluate cross-lingual transfer in the allCPT mT5 setting, where all five Turkic languages are represented in Latin script. Using the recovery rate defined in Section~\ref{par:evaluation-protocol}, we construct transfer matrices where rows correspond to transfer sources and columns correspond to transfer targets. These matrices answer RQ1 by showing how transfer dynamics are shaped by the relationship between the transfer source and the transfer target.

\begin{table*}[!t]
\centering
\tiny
\setlength{\tabcolsep}{1.0pt}
\renewcommand{\arraystretch}{0.72}

\resizebox{0.98\textwidth}{!}{
\begin{tabular}{@{}c@{\hspace{0.8em}}c@{}}

\begin{tabular}{lcccccc}
\toprule
\multicolumn{7}{c}{\textbf{BLEU recovery rate (\%)}} \\
\midrule
\textbf{FT $\backslash$ Eval} & \textbf{tr} & \textbf{az} & \textbf{uz} & \textbf{kk} & \textbf{ky} & \textbf{Avg.} \\
\midrule
tr & \cellcolor{gray!15}-- & \cellcolor{green!45}57.19 & \cellcolor{green!15}35.03 & \cellcolor{red!20}26.59 & \cellcolor{green!15}38.64 & \textbf{39.36} \\
az & \cellcolor{green!25}41.78 & \cellcolor{gray!15}-- & \cellcolor{green!25}40.82 & \cellcolor{green!5}30.65 & \cellcolor{green!25}43.15 & \textbf{39.10} \\
uz & \cellcolor{green!5}33.05 & \cellcolor{green!25}44.59 & \cellcolor{gray!15}-- & \cellcolor{green!15}37.20 & \cellcolor{green!25}42.48 & \textbf{39.33} \\
kk & \cellcolor{green!15}37.15 & \cellcolor{green!25}42.77 & \cellcolor{green!25}39.98 & \cellcolor{gray!15}-- & \cellcolor{green!35}48.99 & \textbf{42.22} \\
ky & \cellcolor{green!5}32.54 & \cellcolor{green!15}38.41 & \cellcolor{green!35}41.82 & \cellcolor{green!25}39.32 & \cellcolor{gray!15}-- & \textbf{38.02} \\
en & \cellcolor{gray!25}30.86 & \cellcolor{gray!25}34.16 & \cellcolor{gray!25}28.08 & \cellcolor{gray!25}29.47 & \cellcolor{gray!25}32.29 & \textbf{30.97} \\
\bottomrule
\end{tabular}

&

\begin{tabular}{lcccccc}
\toprule
\multicolumn{7}{c}{\textbf{chrF recovery rate (\%)}} \\
\midrule
\textbf{FT $\backslash$ Eval} & \textbf{tr} & \textbf{az} & \textbf{uz} & \textbf{kk} & \textbf{ky} & \textbf{Avg.} \\
\midrule
tr & \cellcolor{gray!15}-- & \cellcolor{green!45}79.90 & \cellcolor{green!25}70.17 & \cellcolor{red!20}59.61 & \cellcolor{green!15}69.55 & \textbf{69.81} \\
az & \cellcolor{green!25}71.39 & \cellcolor{gray!15}-- & \cellcolor{green!25}73.16 & \cellcolor{green!5}66.08 & \cellcolor{green!25}74.47 & \textbf{71.28} \\
uz & \cellcolor{green!5}65.55 & \cellcolor{green!25}73.60 & \cellcolor{gray!15}-- & \cellcolor{green!25}71.66 & \cellcolor{green!25}73.34 & \textbf{71.04} \\
kk & \cellcolor{green!15}66.31 & \cellcolor{green!25}71.37 & \cellcolor{green!25}73.35 & \cellcolor{gray!15}-- & \cellcolor{green!45}79.87 & \textbf{72.73} \\
ky & \cellcolor{red!8}63.05 & \cellcolor{green!15}69.96 & \cellcolor{green!25}72.42 & \cellcolor{green!15}68.86 & \cellcolor{gray!15}-- & \textbf{68.57} \\
en & \cellcolor{gray!25}63.11 & \cellcolor{gray!25}63.87 & \cellcolor{gray!25}62.65 & \cellcolor{gray!25}63.05 & \cellcolor{gray!25}65.17 & \textbf{63.57} \\
\bottomrule
\end{tabular}

\end{tabular}
}

\caption{allCPT mT5 Latin recovery-rate transfer matrix. Rows indicate transfer sources and columns indicate transfer targets; non-diagonal cells report recovery averaged over translation targets different from both languages. The English row is a non-Turkic baseline, diagonal cells are omitted, and the Avg. column reports row-wise averages.}
\label{tab:mt5-latin-transfer-recovery}
\end{table*}

\paragraph{Closer Turkic pairs show stronger transfer than English.}
Table~\ref{tab:mt5-latin-transfer-recovery} shows that transfer strength differs clearly across transfer source--transfer target pairs. The strongest BLEU recovery appears from Turkish to Azerbaijani, where tr$\rightarrow$az reaches 57.19\%, and from Kazakh to Kyrgyz, where kk$\rightarrow$ky reaches 48.99\%. The same pattern appears in chrF: tr$\rightarrow$az reaches 79.90\%, while kk$\rightarrow$ky reaches 79.87\%. These pairs also align with the subgroups of the Turkic family: Turkish and Azerbaijani are both Oghuz languages, while Kazakh and Kyrgyz are both Kipchak languages.

English is included as a high-resource non-Turkic transfer source to test whether the strongest transfer results come from general high-resource training or from linguistic relatedness within the Turkic family. In BLEU, English recovers only 28.08--34.16\%, which is below the strongest Turkic transfer pairs. In chrF, English is more stable, reaching 62.65--65.17\%, but it still remains below the intra-Turkic pairs. This suggests that high-resource training alone does not explain the strongest transfer results, and linguistic relatedness within the Turkic family plays an important role.

This is consistent with recent work suggesting that intrinsic language similarity and domain match can support cross-lingual transfer \citep{eronen-etal-2023-improving}, with similar observations reported for Uralic languages \citep{tars-etal-2021-extremely}. While recent studies have established the effectiveness of cross-lingual transfer across the Turkic language family \citep{nguyen-chiang-2017-transfer, yazar2025improving}, our results further provide evidence that the degree of similarity, family subgroup proximity (e.g., within Oghuz or Kipchak branches), is associated with transfer efficiency.

\paragraph{Kazakh is the strongest transfer source on average.}
The Avg. column in Table~\ref{tab:mt5-latin-transfer-recovery} summarizes the overall strength of each language as a transfer source across all transfer targets. In BLEU, Kazakh has the highest average recovery with 42.22\%, followed by Turkish, Uzbek, and Azerbaijani with similar averages around 39\%. The same pattern appears in chrF, where Kazakh again has the highest average recovery with 72.73\%, followed by Azerbaijani with 71.28\% and Uzbek with 71.04\%. 

English has the lowest average in both metrics, with 30.97\% BLEU and 63.57\% chrF, showing that a high-resource non-Turkic transfer source is weaker on average than the Turkic transfer sources. Full recovery matrices for all targets and metrics are provided in Appendix~\ref{app:5-1}.

\subsection{Translation Target Effect (RQ2)}
\begin{figure*}[!tbp]
\centering
\scriptsize
\begin{tikzpicture}
\begin{axis}[
    width=0.84\textwidth,
    height=0.42\textwidth,
    xmin=25, xmax=55,
    xlabel={BLEU recovery rate (\%)},
    ymin=0.5, ymax=6.5,
    y dir=reverse,
    ytick={1,2,3,4,5,6},
    yticklabels={
        {$\mathrm{az}\rightleftarrows\mathrm{ky},\ \mathrm{target}=tr$},
        {$\mathrm{az}\rightleftarrows\mathrm{ky},\ \mathrm{target}=uz$},
        {$\mathrm{az}\rightleftarrows\mathrm{ky},\ \mathrm{target}=kk$},
        {$\mathrm{az}\rightleftarrows\mathrm{kk},\ \mathrm{target}=tr$},
        {$\mathrm{az}\rightleftarrows\mathrm{kk},\ \mathrm{target}=uz$},
        {$\mathrm{az}\rightleftarrows\mathrm{kk},\ \mathrm{target}=ky$}
    },
    yticklabel style={align=right},
    xmajorgrids=true,
    grid style={dashed,gray!30},
    legend style={at={(0.5,-0.16)}, anchor=north, legend columns=2},
]

\addplot[gray!55, thick, forget plot] coordinates {(35.64,1) (52.49,1)};
\addplot[gray!55, thick, forget plot] coordinates {(48.23,2) (34.62,2)};
\addplot[gray!55, thick, forget plot] coordinates {(45.59,3) (28.11,3)};
\addplot[gray!55, thick, forget plot] coordinates {(32.76,4) (51.01,4)};
\addplot[gray!55, thick, forget plot] coordinates {(29.93,5) (37.77,5)};
\addplot[gray!55, thick, forget plot] coordinates {(29.25,6) (39.53,6)};

\addplot[
    only marks,
    mark=*,
    mark size=2.3pt,
    color=blue!70!black,
    mark options={fill=blue!70!black}
] coordinates {
    (35.64,1)
    (48.23,2)
    (45.59,3)
    (32.76,4)
    (29.93,5)
    (29.25,6)
};
\addlegendentry{Right Direction $\rightarrow$}

\addplot[
    only marks,
    mark=square*,
    mark size=2.3pt,
    color=orange!85!black,
    mark options={fill=orange!85!black}
] coordinates {
    (52.49,1)
    (34.62,2)
    (28.11,3)
    (51.01,4)
    (37.77,5)
    (39.53,6)
};
\addlegendentry{Left Direction $\leftarrow$}

\end{axis}
\end{tikzpicture}
\caption{
Paired BLEU recovery-rate comparison for mT5 Latin. Each y-axis label shows a pair of possible transfer source--transfer target directions under a fixed translation target. Direction $\rightarrow$ denotes transfer from the left language to the right language in the y-axis label, while Direction $\leftarrow$ denotes the reverse transfer direction. In the first row, right direction measures az as the transfer source and ky as the transfer target with tr as the translation target The left direction measures the reverse direction, where ky is the transfer source and az is the transfer target.
}
\label{fig:eval-source-target-direction-mt5-latin}
\end{figure*}
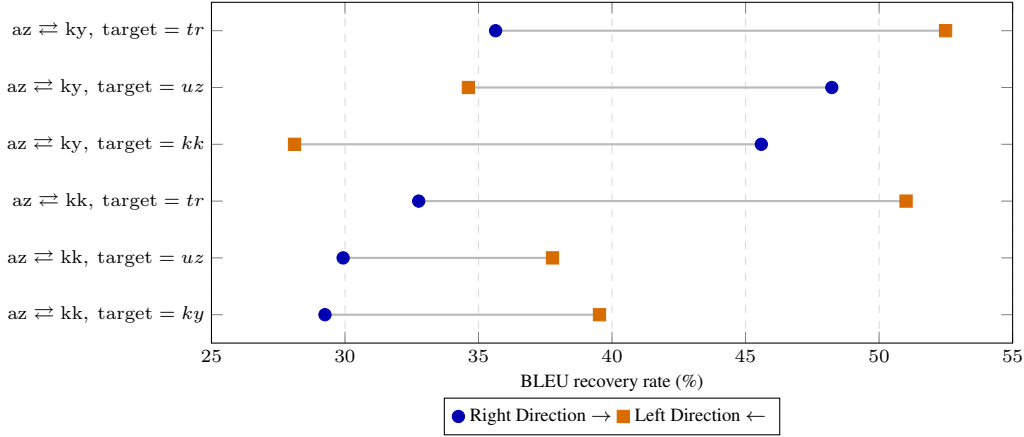

\paragraph{Transfer depends on the translation target.}
Figure~\ref{fig:eval-source-target-direction-mt5-latin} shows that the same transfer source--transfer target pair can produce different recovery rates depending on the translation target. When Azerbaijani is the transfer source and Kyrgyz is the transfer target, recovery changes from 35.64\% with Turkish as the translation target to 48.23\% with Uzbek and 45.59\% with Kazakh. Therefore, transfer cannot be described only as a fixed relationship between two source languages. The translation target also shapes how much knowledge transfers from one source language to another.

\paragraph{Transfer is directional and shaped by transfer-target--translation-target similarity.}
Another interpretation of Figure~\ref{fig:eval-source-target-direction-mt5-latin} is that transfer is directional: transferring from Azerbaijani to Kyrgyz is not equivalent to transferring from Kyrgyz to Azerbaijani under the same translation target. For instance, with Turkish as the translation target, az$\rightarrow$ky recovery is 35.64\%, whereas ky$\rightarrow$az recovery is 52.49\%. A similar asymmetry appears for az$\rightarrow$kk and kk$\rightarrow$az with Turkish as the translation target, where recovery increases from 32.76\% to 51.01\% in the reverse direction. These results suggest that recovery rate tends to be higher when the transfer target and the translation target belong to the same Turkic subgroup, either Oghuz or Kipchak.

A related study suggests that similar target languages can provide stronger positive transfer in one-to-many multilingual MT by using the same translation source for the fine-tuning and evaluation pairs while varying their translation targets \citep{meng-monz-2024-disentangling}. Our setting differs from this design: we keep the translation target fixed and reverse the transfer source--transfer target direction. Therefore, our results provide a complementary view by showing that transfer is also affected by the relationship between the transfer target and the fixed translation target. Additional transfer source--transfer target pairs are provided in Appendix~\ref{app:5-2}

\subsection{Latinization Effects on Transfer (RQ3)} 
\label{sec:latinization-effect}
Latinization is a secondary but important factor because the five languages differ in script usage. Turkish and Azerbaijani are written in Latin script, Kazakh and Kyrgyz are commonly written in Cyrillic, and Uzbek is mixed in practice. Therefore, original-script experiments combine linguistic transfer with orthographic mismatch, while Latinized experiments reduce that script mismatch and make surface overlap easier for the model to use \citep{sun-etal-2022-alternative}.

\begin{table}[!htbp]
\centering
\small
\begin{tabular}{lrrr}
\toprule
\textbf{Transfer} & $\Delta$BLEU (\%) & $\Delta$chrF (\%) & $\Delta$COMET (\%) \\
\midrule
az$\rightarrow$kk & +33.00 & +30.81 & -19.04 \\
az$\rightarrow$ky & +63.64 & +46.08 & -24.49 \\
uz$\rightarrow$kk & +41.03 & +64.06 & -24.91 \\
uz$\rightarrow$ky & +59.63 & +40.79 & -21.27 \\
ky$\rightarrow$kk & -21.01 & -8.20  & -11.06 \\
\bottomrule
\end{tabular}
\caption{
Relative effect of Latinization on allCPT XWMT scores with tr as the translation target. The Transfer column denotes transfer source $\rightarrow$ transfer target. Positive values indicate that Latinization improves the score over the original-script setting, while negative values indicate a decrease.
}
\label{tab:latinization-percent-change-allcpt}
\end{table}

\paragraph{Latinization improves surface metrics for script-mismatched transfer targets.}
Table~\ref{tab:latinization-percent-change-allcpt} reports percent changes from the original-script setting to the Latinized setting. The largest gains appear when the transfer target is Kazakh or Kyrgyz, which are written in Cyrillic in the original-script setting. For example, when Azerbaijani is the transfer source and Kyrgyz is the transfer target, Latinization improves BLEU by 63.64\% and chrF by 46.08\%. Similarly, when Uzbek is the transfer source and Kazakh is the transfer target, Latinization improves BLEU by 41.03\% and chrF by 64.06\%. These gains suggest that Latinization helps the model exploit surface-form similarities that are partly hidden when the transfer source and transfer target are represented in different scripts.

\paragraph{Latinization is not uniformly beneficial across settings and metrics.}
ky$\rightarrow$kk row shows that Latinization does not always improve performance: BLEU decreases by 21.01\% and chrF decreases by 8.20\%. This indicates that Latinization is most useful when it reduces script mismatch between transfer languages. The same rows show negative COMET deltas, which means that better surface-form overlap does not always translate into better semantic metric scores \citep{purkayastha-etal-2023-romanization,soni-bhattacharyya-2024-romantra}. We therefore treat Latinization as a factor that reshapes the matrix, especially for BLEU and chrF, rather than as a uniform improvement across all evaluation criteria. Full Latinized-minus-original matrices are reported in Appendix~\ref{app:5-3}.

\section{Ablation Studies}
\label{sec:correlation-generalization}

The previous section presented the main transfer pattern. In this section we collect correlation-based evidence showing that the transfer patterns we observe are not artifacts of a single dataset, CPT regime, or architecture. These analyses further support the reliability of the observed intra-Turkic transfer relations.

\paragraph{The transfer patterns are mostly stable across different test datasets.}
We first examine the stability of transfer source preferences across XWMT and Tatoeba test datasets. For each fixed evaluation pair $j\rightarrow t$, where $j$ is the transfer target and $t$ is the translation target, we compare all models that were fine-tuned toward the same translation target language $t$ but with different transfer source languages. Each transfer source $i$ defines one candidate donor model, fine-tuned on $i\rightarrow t$ and evaluated on the same pair $j\rightarrow t$. The score obtained on this evaluation pair is treated as the donor score for language $i$.

For example, for the evaluation pair az$\rightarrow$tr, we compare models fine-tuned on az$\rightarrow$tr, uz$\rightarrow$tr, kk$\rightarrow$tr, and ky$\rightarrow$tr, all evaluated on az$\rightarrow$tr. The donor with the highest score is selected as the best donor for that evaluation pair. Repeating this procedure for every evaluation pair gives a donor ranking for each metric and dataset.

We compare the donor rankings obtained from XWMT and Tatoeba using two criteria. \emph{Same best donor} means that the highest-scoring donor is identical on both datasets for the same evaluation pair. \emph{Same full order} means that the complete ranking of donors is identical across the two datasets. Since there are five target languages and four non-target evaluation sources for each target, each metric is compared over 20 donors.

\begin{table}[!htbp]
\centering
\small
\begin{tabular}{llcc}
\toprule
Setting & Metric & Same best donor & Same full order \\
\midrule
Orig.   & BLEU  & 20/20 & 13/20 \\
Orig.   & chrF  & 20/20 & 17/20 \\
Orig.   & COMET &  9/20 &  6/20 \\
\midrule
Latin  & BLEU  & 19/20 & 11/20 \\
Latin  & chrF  & 20/20 & 14/20 \\
Latin  & COMET & 17/20 & 11/20 \\
\bottomrule
\end{tabular}
\caption{
Donor-ranking stability between XWMT and Tatoeba for noCPT mT5.
``Same best donor'' means that the top-ranked fine-tuning donor is the same on both datasets for
a given target and evaluation source. ``Same full order'' means that the complete donor ranking is
identical across the two datasets.
}
\label{tab:benchmark-stability}
\end{table}

Table~\ref{tab:benchmark-stability} shows that BLEU and chrF donor rankings are highly stable across datasets. In the original-script setting, both metrics select the same best donor in all 20 columns; in the Latinized setting, chrF remains perfect and BLEU differs in only one column. Full-order stability is lower, but still strong for BLEU and chrF. COMET is less consistent, especially in the original-script setting, where it preserves the same best donor in only 9/20 columns. Therefore, we treat BLEU and chrF as the main signals for donor-order stability, while COMET serves as complementary evidence. Even though absolute scores differ across XWMT and Tatoeba, the highest-ranked donor usually remains the same for BLEU and chrF. This supports the view that the observed transfer structure is not specific to a single test set, but reflects a more general intra-family transfer pattern among the Turkic languages. Detailed correlation results are provided in Appendix~\ref{app:5-4}.

\paragraph{Transfer scores remain mostly correlated across back-translated training datasets.}
We test the effect of the back-translation model used to create the fine-tuning data on the observed transfer patterns. For this analysis, we compare model results obtained from the original NLLB-generated fine-tuning data with results obtained from a second fine-tuning dataset generated using \texttt{google/madlad400-3b-mt} \citep{kudugunta2023madlad400}.

\begin{table}[!htbp]
\centering
\small
\begin{tabular}{lllcc}
\toprule
Test set & Target & Metric & $r$ & Best donor \\
\midrule
XWMT & az & BLEU & 0.884 & 4/4 \\
XWMT & az & chrF & 0.772 & 4/4 \\
XWMT & kk & BLEU & 0.796 & 4/4 \\
XWMT & kk & chrF & 0.843 & 4/4 \\
\midrule
Tatoeba & az & BLEU & 0.740 & 4/4 \\
Tatoeba & az & chrF & 0.500 & 4/4 \\
Tatoeba & kk & BLEU & 0.601 & 4/4 \\
Tatoeba & kk & chrF & 0.744 & 3/4 \\
\bottomrule
\end{tabular}
\caption{
Stability between NLLB-based and MADLAD-based fine-tuning data for translation targets az and kk.
$r$ denotes Pearson correlation computed over all donor$\times$evaluation cells for each target-specific matrix.
Best donor reports how often the same top-scoring transfer source is selected in both settings.
}
\label{tab:bt-model-stability}
\end{table}

Table~\ref{tab:bt-model-stability} shows positive Pearson correlations between NLLB-based and MADLAD-based fine-tuning matrices in all BLEU and chrF settings. The correlations are strongest on XWMT, ranging from 0.772 to 0.884, which suggests that the relative transfer scores are largely preserved when the back-translation model changes. The correlations on Tatoeba are lower, especially for chrF with Azerbaijani as the translation target, but they remain positive. This indicates that the exact score distribution is affected by the back-translation model and evaluation set, while the overall transfer pattern remains partially consistent. The best-donor agreement gives a similar but coarser signal: the top donor is preserved in almost all cases, with the only one exception. These results further support the generalizability of our findings, while future validation on human-translated data from broader domains would provide an even stronger basis for generalization.

\paragraph{The transfer patterns are largely similar with or without CPT.}
We then examine how continual pretraining affects the structure of cross-source transfer. If CPT substantially changed the transfer behavior, we would expect the noCPT and allCPT settings to produce very different off-diagonal recovery patterns. Instead, Table~\ref{tab:cpt-topology} shows high off-diagonal correlations between the noCPT and allCPT settings.

\begin{table}[!htbp]
\centering
\small
\begin{tabular}{llcc}
\toprule
Comparison & Metric & $\boldsymbol{r_{\mathrm{off}}}$ & Best donor \\
\midrule
allCPT--noCPT & BLEU      & 0.967 & 16/20 \\
allCPT--noCPT & chrF      & 0.915 & 17/20 \\
allCPT--noCPT & COMET     & 0.964 & 11/20 \\
allCPT--noCPT & COMETKiwi & 0.977 & 15/20 \\
\bottomrule
\end{tabular}
\caption{
Stability of cross-source transfer patterns on Latinized XWMT. $r_{\mathrm{off}}$ denotes off-diagonal Pearson correlation and best donor denotes same best donor. High off-diagonal correlation means that the relative pattern of zero-shot transfer is largely preserved.
}
\label{tab:cpt-topology}
\end{table}

The high off-diagonal correlations in Table~\ref{tab:cpt-topology} suggest that CPT does not create a completely new transfer map. For example, allCPT--noCPT reaches $r_{\mathrm{off}}=0.967$ for BLEU and $0.915$ for chrF. This means that CPT can change absolute scores, but the relative organization of transfer remains largely stable. Detailed allCPT-minus-noCPT matrices are provided in Appendix~\ref{app:5-6}.

\paragraph{The transfer patterns are stable across encoder-decoder and decoder-only architectures.}
Finally, we compare the Latinized mT5 matrix with an auxiliary Qwen3 0.6B experiment. This tests whether the raw transfer structure is specific to an encoder--decoder model or whether a similar pattern also appears in a decoder-only model.

\begin{table}[!htbp]
\centering
\small
\begin{tabular}{llcc}
\toprule
Level & Metric & $r$ & $\rho$ \\
\midrule
All raw cells (80)      & BLEU & 0.953 & 0.897 \\
All raw cells (80)      & chrF & 0.929 & 0.896 \\
Off-diagonal raw (60)   & BLEU & 0.883 & 0.784 \\
Off-diagonal raw (60)   & chrF & 0.820 & 0.775 \\
Recovery rate (20)    & BLEU & 0.649 & 0.522 \\
Recovery rate (20)    & chrF & 0.529 & 0.439 \\
\bottomrule
\end{tabular}
\caption{
mT5 Latin vs Qwen3 Latin correlations on Latinized XWMT. $r$ and $\rho$ denote Pearson and Spearman correlation coefficients, respectively.
Raw donor$\times$evaluation matrices are strongly aligned; recovery summaries are
less aligned but still moderately correlated.
}
\label{tab:qwen-corr}
\end{table}

Table~\ref{tab:qwen-corr} shows strong alignment between mT5 and Qwen3 on raw transfer matrices. For all raw cells, Pearson correlation reaches 0.953 for BLEU and 0.929 for chrF; for off-diagonal raw cells, it remains 0.883 for BLEU and 0.820 for chrF. Recovery rate correlations are weaker but still positive. This suggests that architecture changes may rescale or smooth the transfer signal, but they do not erase the family-internal structure observed in the main mT5 matrices. Full Qwen3 recovery tables are reported in Appendix~\ref{app:5-9}.

\section{Conclusion}

This paper examines cross-lingual transfer among Turkish, Azerbaijani, Uzbek, Kazakh, and Kyrgyz through fixed-target transfer matrices. The results show that transfer within the Turkic family depends on both the transfer direction and the translation target. The strongest recovery patterns are concentrated around the Turkish--Azerbaijani and Kazakh--Kyrgyz blocks. At the same time, the same transfer source does not behave uniformly for every transfer target.

Latinization provides a second source of evidence for this interpretation. In script-mismatched directions, especially those involving Kazakh and Kyrgyz, Latinization often improves BLEU and chrF, while these gains are not always mirrored by COMET. This indicates that surface-level improvements and semantic-quality estimates can diverge. Therefore, Latinization should be interpreted as a factor that reshapes transfer patterns rather than as a uniform improvement strategy.

The stability analyses further support the reliability of the observed structure. Donor preferences are highly consistent between XWMT and Tatoeba for BLEU and chrF, and the off-diagonal correlations across noCPT and allCPT show that CPT mostly preserves the transfer pattern even when absolute scores change. The auxiliary Qwen3 experiment leads to a similar conclusion: raw donor--evaluation matrices are strongly correlated with the mT5 results, although recovery summaries are less aligned. Overall, these findings suggest that cross-lingual transfer is a multi-factor phenomenon shaped by the transfer source, transfer target, translation target, script representation, and model setting.

\section*{Limitations}

The main limitation of this study is the use of mostly back-translated bilingual pairs for fine-tuning. This choice was necessary because high-quality direct parallel data between low-resource Turkic languages is limited, and manually creating a balanced human-translated dataset was outside the scope of this work. However, synthetic data can introduce artifacts from the back-translation model and may affect the absolute scores reported in the transfer matrices. For this reason, our claims focus on relative donor preferences, matrix structure, and stability under a fixed experimental pipeline, rather than on absolute translation quality.

A second limitation is that the analysis is restricted to five Turkic languages. The observed correlations suggest that the transfer structure is systematic within this family, but the same conclusions may not directly generalize to other language families with different typological, script, or resource profiles. Repeating the same fixed-target analysis on other low-resource families and on more human-translated data would be an important direction for future work.

\section*{Ethical Considerations}

This work aims to support research on low-resource Turkic machine translation, but the resulting models should not be treated as production-ready systems. Because much of the fine-tuning data is synthetic or automatically mined, translations may contain errors, artifacts, or biases inherited from the back-translation and filtering pipeline. These risks are especially important in high-stakes domains such as legal, medical, or governmental translation.

We use publicly available or automatically generated data and do not intentionally include private or personally identifiable information. However, web-derived corpora may still contain noisy or biased content. We therefore recommend that any released models or datasets be used primarily for research, accompanied by documentation of their construction process and limitations, and carefully evaluated before deployment, especially for minority-language communities where translation errors or uneven data coverage may have disproportionate effects.

\paragraph{Licensing and Copyright.}
For any released data, models, or scripts, we will preserve the required attribution information and release only materials that can be redistributed under the corresponding source licenses. Our code and model releases will be distributed under the Apache License 2.0, while data releases will follow the licenses of the original sources. When a source does not permit redistribution of the original text, we will provide the processing scripts or derived metadata instead of the restricted content. This is intended to support reproducibility while respecting the intellectual property rights of the original data providers and authors.

\paragraph{Reproducibility.}
To support reproducibility, we release the dataset construction scripts, Latinization rules, fine-tuning and evaluation scripts, and result matrices used in our experiments. The repository also includes the configuration files and instructions needed to reproduce the main transfer-matrix results. All reported findings should be interpreted with respect to our specific experimental setup, including the selected models, preprocessing pipeline, decoding parameters, evaluation datasets, and metric implementations.

\paragraph{Use of Generative AI.}
Generative AI tools were used only to assist with language editing and improving the clarity of the manuscript. All scientific contributions, experimental design, data construction decisions, result analysis, and interpretations were conducted and verified by the authors.

\section*{Acknowledgments}

We thank Google Cloud, Google Cloud Academic Credit Program, and TRUBA (Turkish Science e-Infrastructure) for providing the compute resources that made this study possible.

\appendix
\setlength{\parskip}{0pt}

\section{Appendix}

\subsection{Dataset Creation Details}
\label{app:dataset-details}

\subsubsection{Monolingual Corpus}

Large-scale monolingual corpora were constructed for five Turkic languages (Turkish, Azerbaijani, Kazakh, Kyrgyz, and Uzbek) by collecting data from publicly available resources, including Wikipedia, CC-100 \cite{conneau-etal-2020-unsupervised}, Leipzig Corpora \cite{goldhahn-etal-2012-building}, OSCAR \cite{abadji-etal-2022-towards}, mC4 \cite{xue-etal-2021-mt5}, MADLAD-400 \cite{kudugunta2023madlad400}, and HPLT \cite{aulamo-etal-2023-hplt}. Additional language-specific datasets were incorporated where available. The collected raw text was cleaned through line-level filtering (removal of short, numeric, or noisy lines), paragraph reconstruction, and MD5-based deduplication. The final datasets were stored in JSONL format for each language. The amount of data collected for each language is shown in Table~\ref{tab:mono-data}. For continual pretraining (CPT), the data was segmented into approximately 480-token chunks using the \texttt{google/mt5-small} tokenizer \cite{xue-etal-2021-mt5}, while preserving sentence boundaries. Cyrillic-based languages were transliterated into their official Latin alphabet versions.

\begin{table}[!htbp]
\centering
\begin{tabular}{lccc}
\toprule
\textbf{Language} & \textbf{Number of Chunks} & \textbf{File Size} \\
\midrule
Turkish   & 69,848,795 & 102GB \\
Azerbaijani  & 16,960,000 & 25GB \\
Kazakh  & 6,638,029 & 27GB \\
Kyrgyz & 3,710,962 & 13GB \\
Uzbek  & 8,265,191 & 17GB \\
\bottomrule
\end{tabular}
\caption{Distribution of Monolingual Data}
\label{tab:mono-data}
\end{table}

\subsubsection{Bilingual Corpus}

\begin{figure}[!htbp]
     \centering
     
     \begin{subfigure}[b]{0.48\textwidth}
         \centering
         \includegraphics[width=\linewidth]{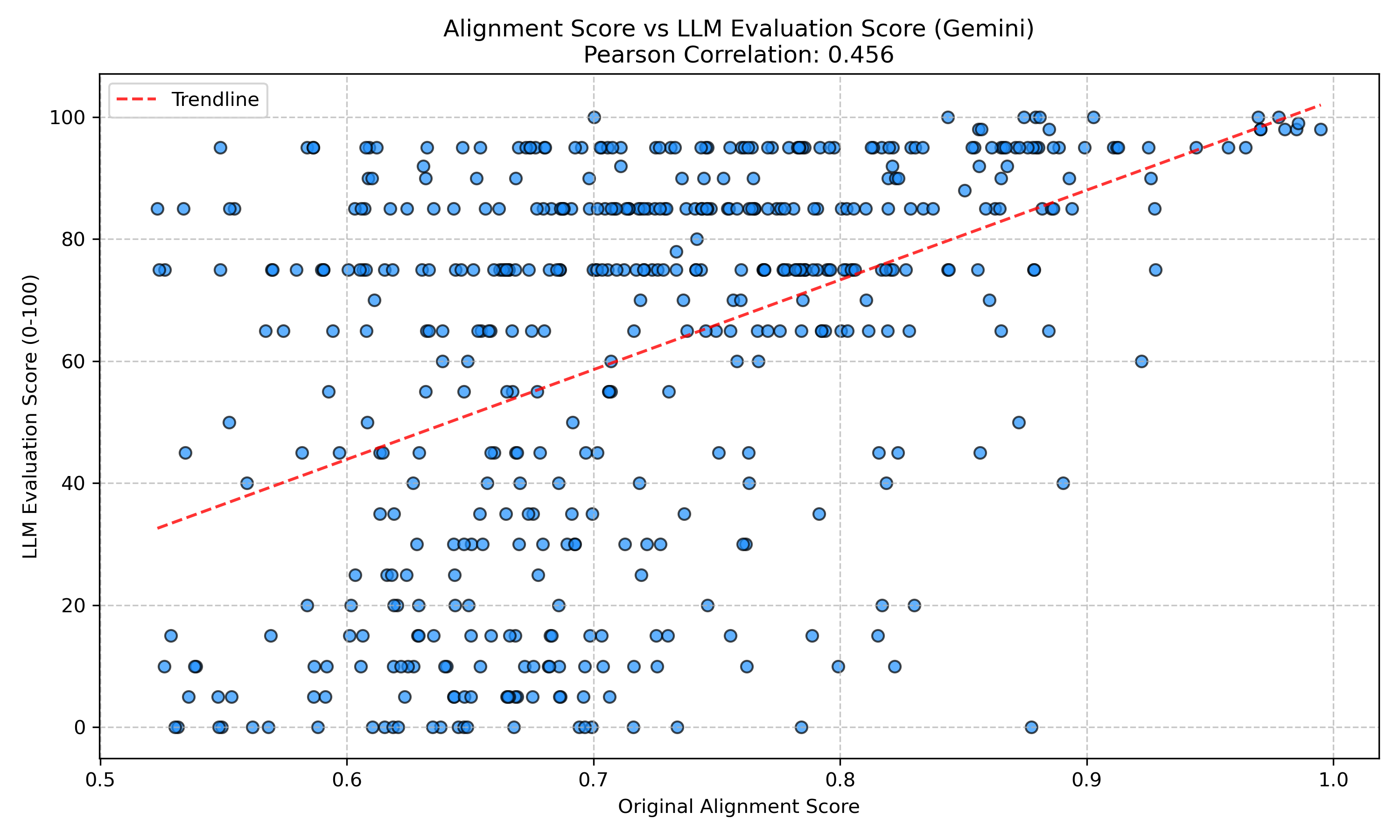}
         \caption{LaBSE Score vs. LLM Score}
         \label{fig:score_all}
     \end{subfigure}
     \hfill 
     
     \begin{subfigure}[b]{0.48\textwidth}
         \centering
         \includegraphics[width=\linewidth]{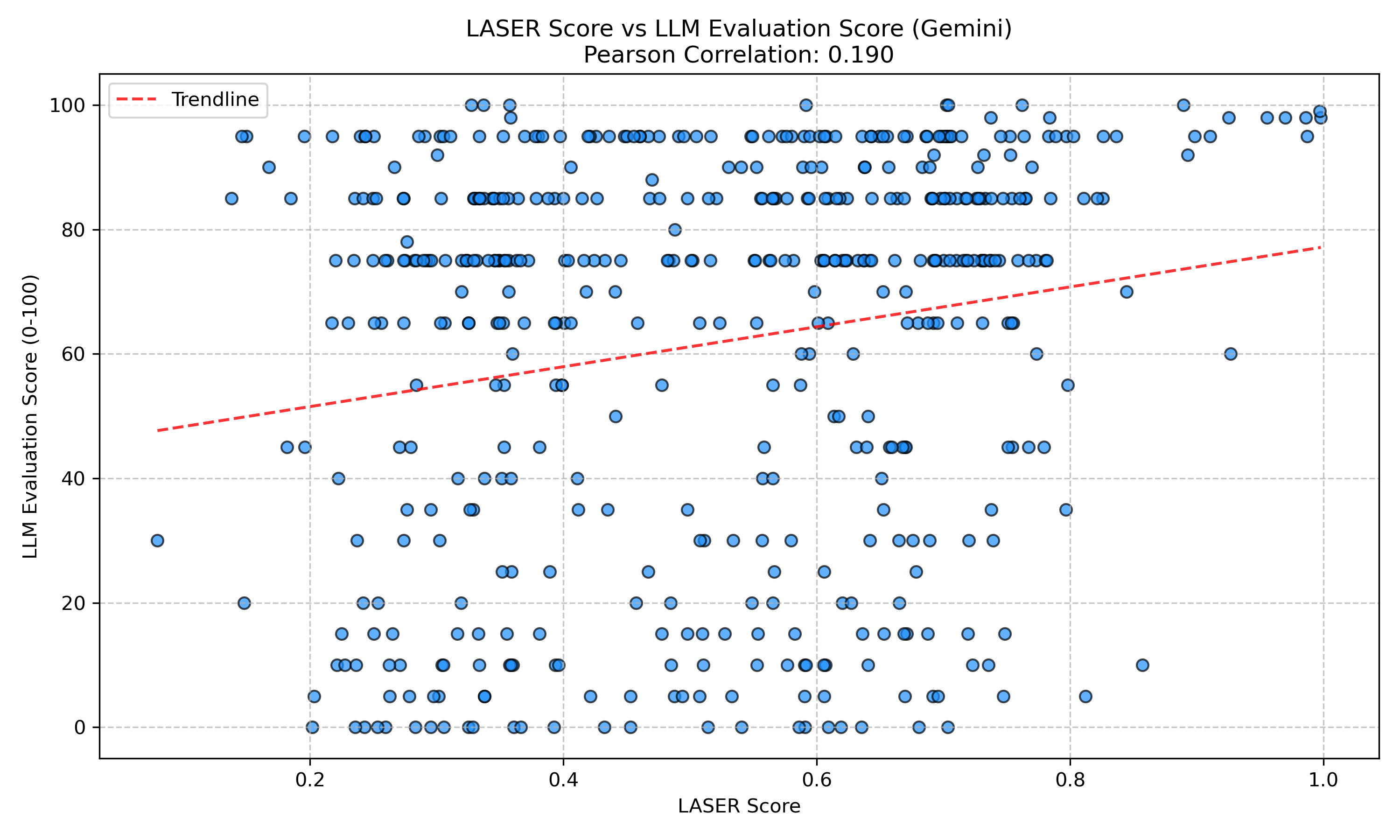}
         \caption{LASER Score vs. LLM Score}
         \label{fig:score_laser}
     \end{subfigure}
     
     \caption{Comparison of Dataset Filtering Metrics}
     \label{fig:comparison_main}
\end{figure}

Parallel corpora were obtained from three main sources: pivot-based extraction on publicly available datasets, back-translation, and open-source books available in multiple Turkic languages. Direct and pivot-based bilingual pairs were obtained from KazParC \cite{yeshpanov-etal-2024-kazparc}, NTREX \cite{federmann-etal-2022-ntrex}, and FLORES+ \cite{gordeev-etal-2024-flores}, OPUS \cite{tiedemann-2012-parallel} including corpora such as OpenSubtitles, TED2020, and CCAligned. All parallel data underwent multi-stage filtering, including language detection via FastText \cite{bojanowski2017enriching}, script normalization, removal of noisy or misaligned pairs, length ratio constraints, and deduplication. Since the collected data were highly imbalanced and contained few direct pairs between low-resource Turkic languages, synthetic parallel data was generated via back-translation using \texttt{facebook/nllb-200-distilled-600M} \cite{nllb2022}, with quality filtering based on LaBSE \cite{feng-etal-2022-language} similarity scores and LLM-as-a-judge evaluation method. Additionally, parallel data was extracted from multilingual book translations using LaBSE-based bitext mining, applying both paragraph- and sentence-level alignment with cosine similarity thresholds. Comparative evaluation against LASER \cite{artetxe-schwenk-2019-laser} using an LLM-based scoring framework \cite{kocmi-federmann-2023-large} demonstrated stronger correlation for LaBSE, which was therefore used for semantic filtering (Figure~\ref{fig:comparison_main}). These results also correlate with the work by \citet{chimoto-bassett-2022-low} which stated that LaBSE works better than LASER for low-resource languages with a study on African languages. The prompt given to our judge LLM (Gemini 3.1 Pro) to score the quality of the machine translation is taken from \cite{kocmi-federmann-2023-large} and can be found below. 
\\
\begin{mdframed}[backgroundcolor=gray!5,linecolor=gray!75]
\textbf{LLM Evaluation Prompt}
\textit{
Score the following translation from \{Source Language\} to \{Target Language\} on a continuous scale from 0 to 100, where a score of zero means ``no meaning preserved'' and a score of one hundred means ``perfect meaning and grammar''. \\
\{Source Language\} source: \{Source Sentence\} \\
\{Target Language\} translation: \{Translated Sentence\} \\
Score:
}
\end{mdframed}

\noindent
In this work, we used pairs from publicly available datasets for CPT in transfer-coefficient sampling experiments, and we combined back-translated data with book-aligned pairs to create the FT data for transfer-matrix calculations. We used all aligned book pairs and completed the FT dataset size to 100K pairs for each language using back-translation \citep{de-gibert-etal-2025-scaling}, following the 100K training-size setting used by \citet{eronen-etal-2023-improving}. The number of manually curated book-aligned pairs among Turkic languages is shown in Table~\ref{tab:dist}. For inference / test data we used a subset of \texttt{turkic-interlingua/turkic\_xwmt} test dataset \cite{mirzakhalov-etal-2021-large} which contains 400 pairs between each language pairs and as the secondary test set for correlation experiments we used a subset of Tatoeba dataset \citep{tiedemann-2020-tatoeba} which also contains 400 pairs for each language pair.

\begin{table}[!htbp]
\centering
\begin{tabular}{lc}
\toprule
\textbf{Language Pair (Source-Target)} & \textbf{Total} \\
\midrule
tr-kk & 7605 \\
tr-az & 6982 \\
az-uz & 2104 \\
tr-uz & 1999 \\
kk-ky & 275 \\
tr-ky & 255 \\
\midrule
\textbf{Total} & \textbf{19220} \\
\bottomrule
\end{tabular}
\caption{Distribution of Book Pairs between Languages}
\label{tab:dist}
\end{table}

\subsection{Implementation Details}
\label{app:implementation-details}

\subsubsection{CPT and FT Scripts for mT5-small}
The experiments use \texttt{mT5-small} (300M parameter) as the base sequence-to-sequence model. Continual pretraining follows the T5 span-corruption objective: each monolingual chunk is tokenized, random spans are replaced with sentinel tokens, and the target sequence consists of the removed spans. The same CPT hyperparameters are used across all CPT scripts, as reported in Table~\ref{tab:appendix-cpt-hparams}.

\begin{table*}[!tbp]
\centering
\small
\begin{tabular*}{\textwidth}{@{\extracolsep{\fill}}ll@{}}
\toprule
\textbf{CPT setting} & \textbf{Value} \\
\midrule
Base architecture & \texttt{google/mt5-small} / local \texttt{mT5-small} copy \\
Objective & T5 span corruption \\
Maximum examples & 1,000,000 per language for allCPT \\
Maximum token length & 512 \\
Noise density & 0.15 \\
Mean noise span length & 3.0 \\
Training epochs & 1 \\
Batch size & 32 \\
Gradient accumulation & 1 \\
Learning rate & $5\times10^{-5}$ \\
Weight decay & 0.01 \\
Warmup ratio & 0.03 \\
Optimizer & \texttt{adamw\_torch} \\
Precision & bf16 \\
Save policy & step-based saving with \texttt{save\_total\_limit=1} \\
Seed & 42 \\
\bottomrule
\end{tabular*}
\caption{Shared CPT hyperparameters used across the CPT scripts.}
\label{tab:appendix-cpt-hparams}
\end{table*}

Fine-tuning uses a fixed-translation-target transfer design. For each translation target language $t$, the model is fine-tuned on one translation source--target pair $i\rightarrow t$ and evaluated on all source languages $j\rightarrow t$ for transfer target $j\neq t$. Inputs are formatted as explicit source and target tags followed by the source sentence. The same fine-tuning and decoding hyperparameters are used across the FT scripts, and these shared settings are reported in Table~\ref{tab:appendix-ft-hparams}. Because GPU resources were limited, we did not perform an exhaustive hyperparameter search and instead adopted the learning-rate setting from \citet{yu-etal-2023-simple}.

For all CPT and FT experiments, a single Nvidia A100 80GB GPU is used via Google Cloud Console Virtual Machine Instances. Fine-tuning mT5-small model with a single translation source--target pair $i\rightarrow t$ took 1 hour 40 minutes to 2 hours and CPT with 5M total chunks each of approximately 512 tokens took 5 to 6 hours.

\begin{table*}[!tbp]
\centering
\small
\begin{tabular*}{\textwidth}{@{\extracolsep{\fill}}ll@{}}
\toprule
\textbf{Fine-tuning/evaluation setting} & \textbf{Value} \\
\midrule
Languages & \texttt{tr}, \texttt{az}, \texttt{uz}, \texttt{kk}, \texttt{ky} \\
Input format & \texttt{<src> <tgt> source sentence} \\
Training epochs & 5.0 \\
Train batch size & 8 \\
Eval batch size & 8 \\
Gradient accumulation & 4 \\
Learning rate & $3\times10^{-4}$ \\
Weight decay & 0.01 \\
Warmup ratio & 0.1 \\
Scheduler & linear \\
Validation split & 0.05 \\
Max source / target length & 256 / 256 \\
Generation & beam size 5, maximum length 256, early stopping \\
Maximum test samples & 400 \\
Metrics & sacreBLEU, chrF, COMET, COMETKiwi \\
COMET models & \texttt{Unbabel/wmt22-comet-da}, \texttt{Unbabel/wmt22-cometkiwi-da} \\
Seed & 42 \\
\bottomrule
\end{tabular*}
\caption{Shared fine-tuning and evaluation hyperparameters used in the transfer-matrix scripts.}
\label{tab:appendix-ft-hparams}
\end{table*}

\subsubsection{FT and Evaluation Scripts for Qwen3 0.6B}
\label{app:qwen3-implementation}

To test whether the transfer structure observed with mT5 is specific to an encoder--decoder architecture or not, we also run an auxiliary decoder-only experiment with \texttt{Qwen/Qwen3-0.6B}. The Qwen3 experiment follows the same fixed-translation-target transfer logic as the mT5 transfer-matrix experiments: for each pivot translation target language $t$, the model is fine-tuned on one donor direction (both transfer and translation source) $i\rightarrow t$ and then evaluated on all test directions $j\rightarrow t$ where transfer target $j\neq t$. The resulting matrices use fine-tuning donor (transfer source) languages as rows and evaluation source (transfer target) languages as columns.

Because Qwen3 is a causal language model, the translation input and target are concatenated into a single sequence. Each training example is formatted as a natural-language translation prompt followed by the reference translation:
\begin{quote}
\small
\texttt{Translate from SourceLanguage to TargetLanguage: source sentence \textbackslash n}\\
\texttt{target sentence <eos>}
\end{quote}
The loss is computed only on the target/completion tokens. Prompt tokens are masked with \texttt{-100} in the label sequence, so that they do not contribute to the training objective. During generation, the same prompt format is used, and the prompt portion is removed from the decoded output before scoring. Implementation details can be seen on Table~\ref{tab:appendix-qwen3-ft-hparams}. Similar to mT5, we did not perform hyperparameter search and used the Qwen3 fine-tuning learning-rate from \citet{luo2025beyond}.

\begin{table*}[!hbp]
\centering
\small
\begin{tabular}{ll}
\toprule
\textbf{Qwen3 FT/Eval setting} & \textbf{Value} \\
\midrule
Base model & \texttt{Qwen/Qwen3-0.6B} \\
Model class & \texttt{AutoModelForCausalLM} \\
Tokenizer class & \texttt{AutoTokenizer} \\
Training framework & Hugging Face \texttt{Trainer} / \texttt{TrainingArguments} \\
Fine-tuning design & Fixed-translation-target translation source fine-tuning, evaluated for all transfer targets \\
Prompt format & \texttt{Translate from \{SourceLanguage\} to \{TargetLanguage\}: source\textbackslash n} \\
Training sequence & Prompt + target + \texttt{eos} \\
Loss masking & Prompt labels set to \texttt{-100}; loss computed only on target tokens \\
Training epochs & 3.0 \\
Train batch size & 4 \\
Eval batch size & 4 \\
Gradient accumulation & 8 \\
Effective batch size & 32 \\
Learning rate & $2\times10^{-5}$ \\
Weight decay & 0.01 \\
Warmup ratio & 0.1 \\
Scheduler & cosine \\
Max gradient norm & 1.0 \\
Validation split & 0.05 \\
Max source / target length & 256 / 256 \\
Maximum generated tokens & 256 \\
Generation & beam size 5, early stopping, no sampling \\
Maximum test samples & 400 per evaluation pair \\
Precision & fp16 disabled; bf16 disabled \\
Metrics & sacreBLEU, chrF \\
Seed & 42 \\
\bottomrule
\end{tabular}
\caption{Qwen3 fine-tuning and evaluation hyperparameters used in the auxiliary decoder-only transfer experiment.}
\label{tab:appendix-qwen3-ft-hparams}
\end{table*}

For evaluation, the fine-tuned Qwen3 model is loaded once for each translation source--target run and reused across all evaluation sources (transfer targets) for that fixed translation target. The tokenizer is switched to left padding during generation, which is required for batched causal-LM decoding. Predictions are normalized for whitespace and punctuation before scoring with sacreBLEU and chrF. For each run, the script stores per-pair scores, transfer matrices, logs, and generated predictions in JSONL format. For Qwen3 FT and evaluation experiments dual Nvidia P100 16GB GPU is used via TRUBA ARF barbun-cuda HPC servers. Fine-tuning Qwen3 0.6B model with a single translation source--target pair $i\rightarrow t$ and obtaining the evaluation results took approximately 14 hours on this GPU setup.

\subsubsection{Examples From Training and Test Datasets}
The following examples illustrate the JSONL formats used by the scripts. Training files contain explicit source and target language fields, source and target texts, and metadata about pivoting or synthetic generation. Test files follow the common \texttt{translation} dictionary format used by XWMT/Tatoeba-style datasets.
\\
\begin{lstlisting}[language={}]
{"src_lang":"az","tgt_lang":"tr","src":"Cengiz Hanin hokmranligi dovrunde 1206-1227-ci iller arasinda Simali Cinde Bati Xia ve Jin Hanedani; Turkistanda Kara Hitay, Maveraunnehir; Iranda Harezm, Horasan ve Harezmsahlar, Kafkasyada Gurculer, Dest-i Kipcakdaki Rus Knezleri, Kipcaklar ve Idil Bolqarlar uzerinde seferler", "tgt":"Cengiz Han, hukumdarligi doneminde, 1206-1227 arasinda, Kuzey Cin'deki Bati Xia ve Jin Hanedani; Turkistan'daki Kara Hitay, Maveraunnehir; Harezm, Horasan ve Iran'daki Harezmsahlar, Kafkasya'daki Gurculer, Dest-i Kipcak'taki Rus Knezlikleri, Kipcaklar ile Idil Bulgarlari uzerine seferler yapti ve imparatorlugu doneminde gerceklestirdigi hicbir savasi kaybetmedi.", "idx":0, "pivot":"tr", "nllb_model":"facebook/nllb-200-distilled-600M", "second":"none"}
\end{lstlisting}

\begin{lstlisting}[language={}]
{"translation":{"tr":"Kizim yurt disinda okuyor.","ky":"Kizim cet olkodo okuyt."}}
{"translation":{"tr":"Artik kime inanacagimi bilmiyorum.","ky":"Emi kimge isenerimdi bilbeym."}}
{"translation":{"tr":"Buyuk bir arabam var.","ky":"Menin con masinam bar."}}
\end{lstlisting}

\subsubsection{Example Model Outputs}
Table~\ref{tab:appendix-model-output-examples} shows example outputs from the allCPT Latinized XWMT evaluation directory. All examples use Turkish as the fixed translation target and the model fine-tuned on \texttt{uz}$\rightarrow$\texttt{tr}; only the evaluation source (transfer target) changes.

\begin{table*}[!bp]
\centering
\scriptsize
\setlength{\tabcolsep}{3pt}
\begin{tabular}{p{0.08\textwidth}p{0.25\textwidth}p{0.30\textwidth}p{0.30\textwidth}}
\toprule
\textbf{Eval pair} & \textbf{Source} & \textbf{Reference} & \textbf{Prediction} \\
\midrule
az$\rightarrow$tr & IB rəsmiləri əlavə olaraq qeyd edirlər ki, niyyət terrorçulara sular vasitəsilə sızmaq da olabilər. & Niyet, teroristlerin sudan sizdirilmasi olacagi, IB yetkilileri tarafindan ek olarak not edildi. & IB resmileri iptal edilebilir ki , niyet terrorçulara sular vasitəsiyle sızmaq da olabilər. \\
kk$\rightarrow$tr & Maqsat - lañkesterge su arqyly enu , dep qosty aqparattyq agenttik. & Niyet, teroristlerin sudan sizdirilmasi olacagi, IB yetkilileri tarafindan ek olarak not edildi. & İlk olarak lañkesterge su aracılığıyla enu , olarak kurduğu amatör agenttik. \\
ky$\rightarrow$tr & "Indiyanın çalgındoo byurosu bildirgendey, terroristter deñizge kirüü ıkmaların üyrötüp jatışat. & Niyet, teroristlerin sudan sizdirilmasi olacagi, IB yetkilileri tarafindan ek olarak not edildi. & Ülkede ikinci olarak, İran'ın çalgındoo bürosu bildirdiği , terroristler denize girdik ıkmaların üyrütüp gitti. \\
uz$\rightarrow$tr & Shuningdek , maqsad terroristlar orasiga suv orqali kirib borish deya Axborot agentligi rasmiylari qo’shimcha qiladi. & Niyet, teroristlerin sudan sizdirilmasi olacagi, IB yetkilileri tarafindan ek olarak not edildi. & Ayrıca amaç terroristler arasında su aracılığıyla erişilmesi olduğuna Güvenlik Enstitüsü resmî resmi sayısı dahil olmak üzere işaret eder. \\
\bottomrule
\end{tabular}
\caption{Example generated predictions from the allCPT Latinized XWMT output directory from uz$\rightarrow$tr FT.}
\label{tab:appendix-model-output-examples}
\end{table*}

\subsubsection{Reproducibility Statement}
Our GitHub repository \footnote{https://github.com/TurkicTransfer/Cross-LingualTransferforMachineTranslationinTurkicLanguages} contains:
\begin{itemize}
    \item Dataset construction scripts;
    \item Latinization scripts and rules;
    \item CPT, fine-tuning and evaluation scripts;
    \item HTML files that contain the full result matrices and comparison summaries.
\end{itemize}
To reproduce the main mT5 transfer matrices, run the CPT checkpoint preparation step, fine-tune one model for each ordered source--target pair, and then run the fixed-target evaluation script with \texttt{MAX\_TEST\_SAMPLES=400}. For COMET and COMETKiwi, the Hugging Face token must allow access to the corresponding \texttt{Unbabel} checkpoints. All runs use seed 42.

\setlength{\LTleft}{0pt}\setlength{\LTright}{0pt}

\subsection{Detailed mT5 Results as Complementary Data}

\subsubsection{Latinized allCPT XWMT Recovery Tables}
\label{app:5-1}

\begin{figure*}[!t]
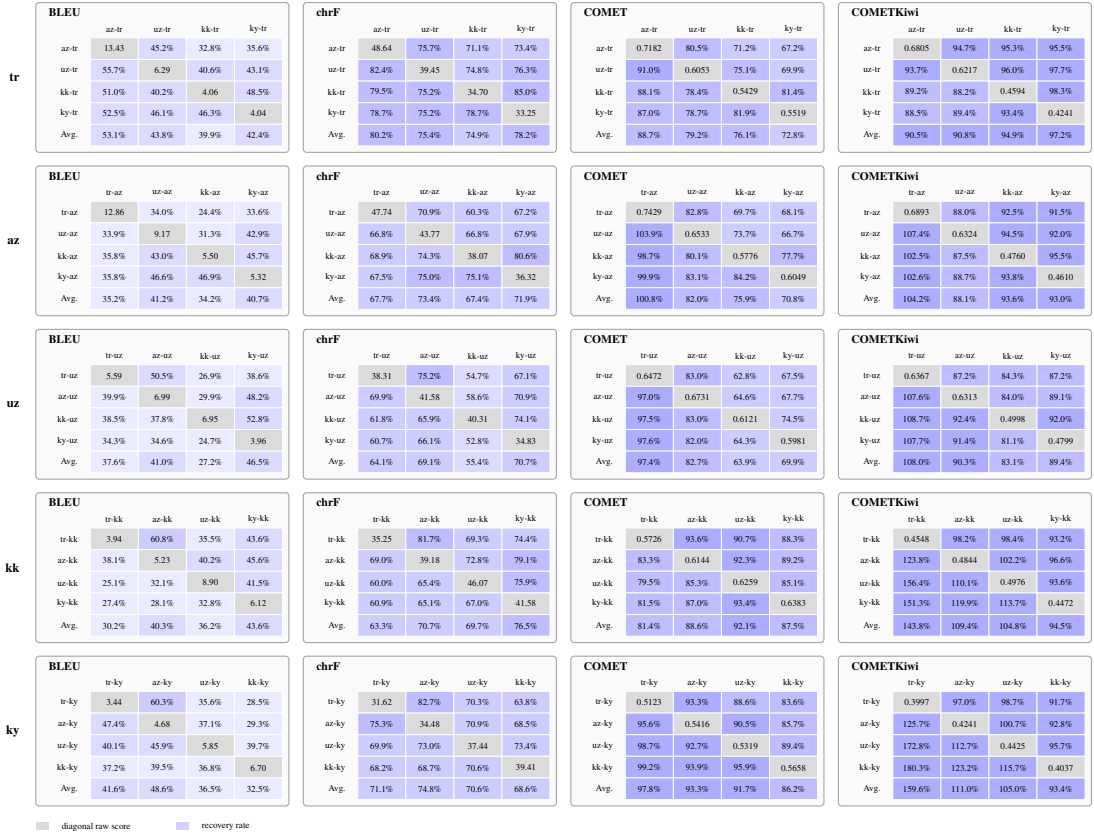

\centering
\begin{adjustbox}{width=\textwidth,max totalheight=0.96\textheight,center}

\end{adjustbox}
\caption{Full Latinized allCPT XWMT recovery matrices. Rows of panels correspond to fixed translation targets and metric columns correspond to BLEU, chrF, COMET, and COMETKiwi. Diagonal cells report raw same-source scores, off-diagonal cells report recovery rates, and the Avg. row reports the average recovery for each fine-tuning source.}
\label{fig:app-allcpt-recovery-exact}
\end{figure*}

Figure~\ref{fig:app-allcpt-recovery-exact} reports the full Latinized allCPT XWMT recovery matrices used to support the main transfer analysis. The rows of panels correspond to fixed translation targets, while the metric columns show BLEU, chrF, COMET, and COMETKiwi. Within each panel, diagonal cells report the raw same-source score for the corresponding fine-tuning/evaluation pair, and off-diagonal cells report the recovery rate obtained when the fine-tuning source differs from the evaluation source. This organization preserves the complete numerical evidence behind the averaged transfer matrix in the main paper while making the target-conditioned structure visible: strong transfer is concentrated around related-language blocks, but the same donor can behave differently depending on the fixed translation target and evaluation source.

\subsubsection{Translation Target Effect}
\label{app:5-2}

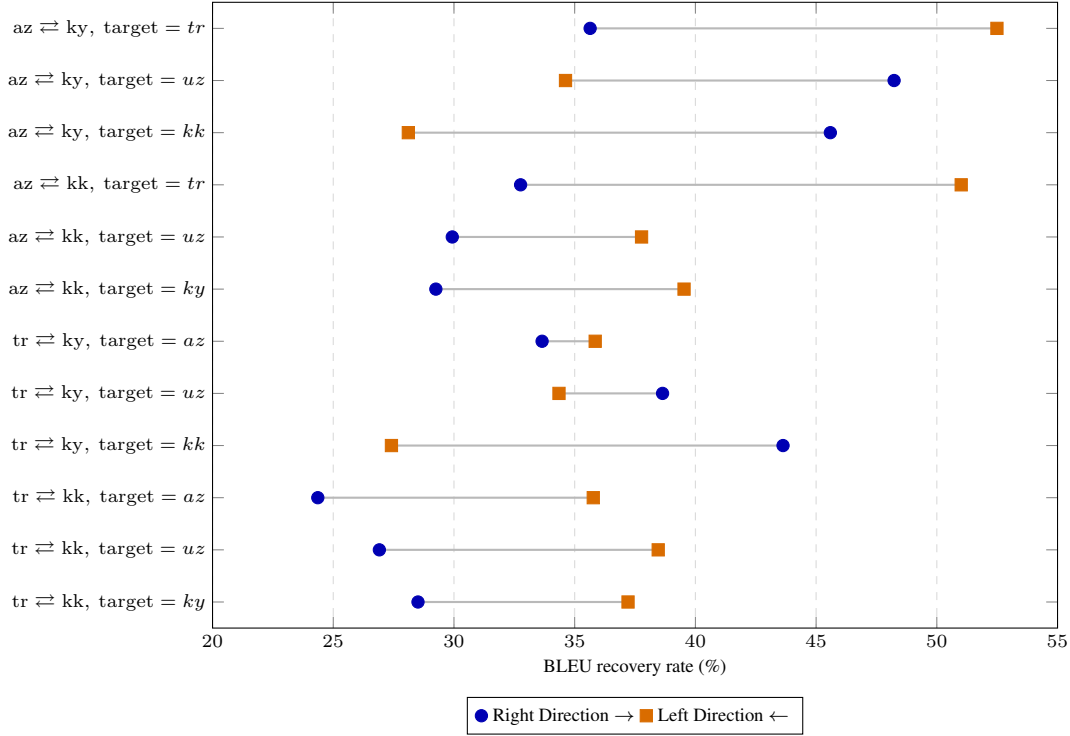
\begin{figure*}[!tbp]
\centering
\scriptsize
\begin{tikzpicture}
\begin{axis}[
    width=0.88\textwidth,
    height=0.68\textwidth,
    xmin=20, xmax=55,
    xlabel={BLEU recovery rate (\%)},
    ymin=0.5, ymax=12.5,
    y dir=reverse,
    ytick={1,2,3,4,5,6,7,8,9,10,11,12},
    yticklabels={
        {$\mathrm{az}\rightleftarrows\mathrm{ky},\ \mathrm{target}=tr$},
        {$\mathrm{az}\rightleftarrows\mathrm{ky},\ \mathrm{target}=uz$},
        {$\mathrm{az}\rightleftarrows\mathrm{ky},\ \mathrm{target}=kk$},
        {$\mathrm{az}\rightleftarrows\mathrm{kk},\ \mathrm{target}=tr$},
        {$\mathrm{az}\rightleftarrows\mathrm{kk},\ \mathrm{target}=uz$},
        {$\mathrm{az}\rightleftarrows\mathrm{kk},\ \mathrm{target}=ky$},
        {$\mathrm{tr}\rightleftarrows\mathrm{ky},\ \mathrm{target}=az$},
        {$\mathrm{tr}\rightleftarrows\mathrm{ky},\ \mathrm{target}=uz$},
        {$\mathrm{tr}\rightleftarrows\mathrm{ky},\ \mathrm{target}=kk$},
        {$\mathrm{tr}\rightleftarrows\mathrm{kk},\ \mathrm{target}=az$},
        {$\mathrm{tr}\rightleftarrows\mathrm{kk},\ \mathrm{target}=uz$},
        {$\mathrm{tr}\rightleftarrows\mathrm{kk},\ \mathrm{target}=ky$}
    },
    yticklabel style={align=right},
    xmajorgrids=true,
    grid style={dashed,gray!30},
    legend style={at={(0.5,-0.11)}, anchor=north, legend columns=2},
]

\addplot[gray!55, thick, forget plot] coordinates {(35.64,1) (52.49,1)};
\addplot[gray!55, thick, forget plot] coordinates {(48.23,2) (34.62,2)};
\addplot[gray!55, thick, forget plot] coordinates {(45.59,3) (28.11,3)};

\addplot[gray!55, thick, forget plot] coordinates {(32.76,4) (51.01,4)};
\addplot[gray!55, thick, forget plot] coordinates {(29.93,5) (37.77,5)};
\addplot[gray!55, thick, forget plot] coordinates {(29.25,6) (39.53,6)};

\addplot[gray!55, thick, forget plot] coordinates {(33.65,7) (35.85,7)};
\addplot[gray!55, thick, forget plot] coordinates {(38.64,8) (34.35,8)};
\addplot[gray!55, thick, forget plot] coordinates {(43.63,9) (27.41,9)};

\addplot[gray!55, thick, forget plot] coordinates {(24.36,10) (35.77,10)};
\addplot[gray!55, thick, forget plot] coordinates {(26.91,11) (38.46,11)};
\addplot[gray!55, thick, forget plot] coordinates {(28.51,12) (37.21,12)};

\addplot[
    only marks,
    mark=*,
    mark size=2.3pt,
    color=blue!70!black,
    mark options={fill=blue!70!black}
] coordinates {
    (35.64,1)
    (48.23,2)
    (45.59,3)
    (32.76,4)
    (29.93,5)
    (29.25,6)
    (33.65,7)
    (38.64,8)
    (43.63,9)
    (24.36,10)
    (26.91,11)
    (28.51,12)
};
\addlegendentry{Right Direction $\rightarrow$}

\addplot[
    only marks,
    mark=square*,
    mark size=2.3pt,
    color=orange!85!black,
    mark options={fill=orange!85!black}
] coordinates {
    (52.49,1)
    (34.62,2)
    (28.11,3)
    (51.01,4)
    (37.77,5)
    (39.53,6)
    (35.85,7)
    (34.35,8)
    (27.41,9)
    (35.77,10)
    (38.46,11)
    (37.21,12)
};
\addlegendentry{Left Direction $\leftarrow$}

\end{axis}
\end{tikzpicture}
\caption{
Extended appendix version of the paired BLEU recovery-rate comparison for mT5 Latin. Each y-axis label shows two opposite transfer source--transfer target directions under a fixed translation target. Direction $\rightarrow$ denotes transfer from the left language to the right language in the y-axis label, while Direction $\leftarrow$ denotes transfer in the reverse direction. 
}
\label{fig:appendix-eval-source-target-direction-mt5-latin}
\end{figure*}

Figure~\ref{fig:appendix-eval-source-target-direction-mt5-latin} extends the target-effect analysis from the main paper by showing additional bidirectional transfer comparisons under different fixed translation targets. Each line connects two opposite transfer directions for the same language pair and translation target. Large gaps between the two markers indicate directional asymmetry, while changes across rows show that the same transfer pair can behave differently when the translation target changes. These patterns support the main claim that cross-lingual transfer among Turkic languages is target-conditioned rather than reducible to a single donor--recipient similarity score.

\subsubsection{Nominal Deltas for Latinized - Original XWMT test dataset Raw Scores on allCPT}
\label{app:5-3}

\begin{figure*}[!t]
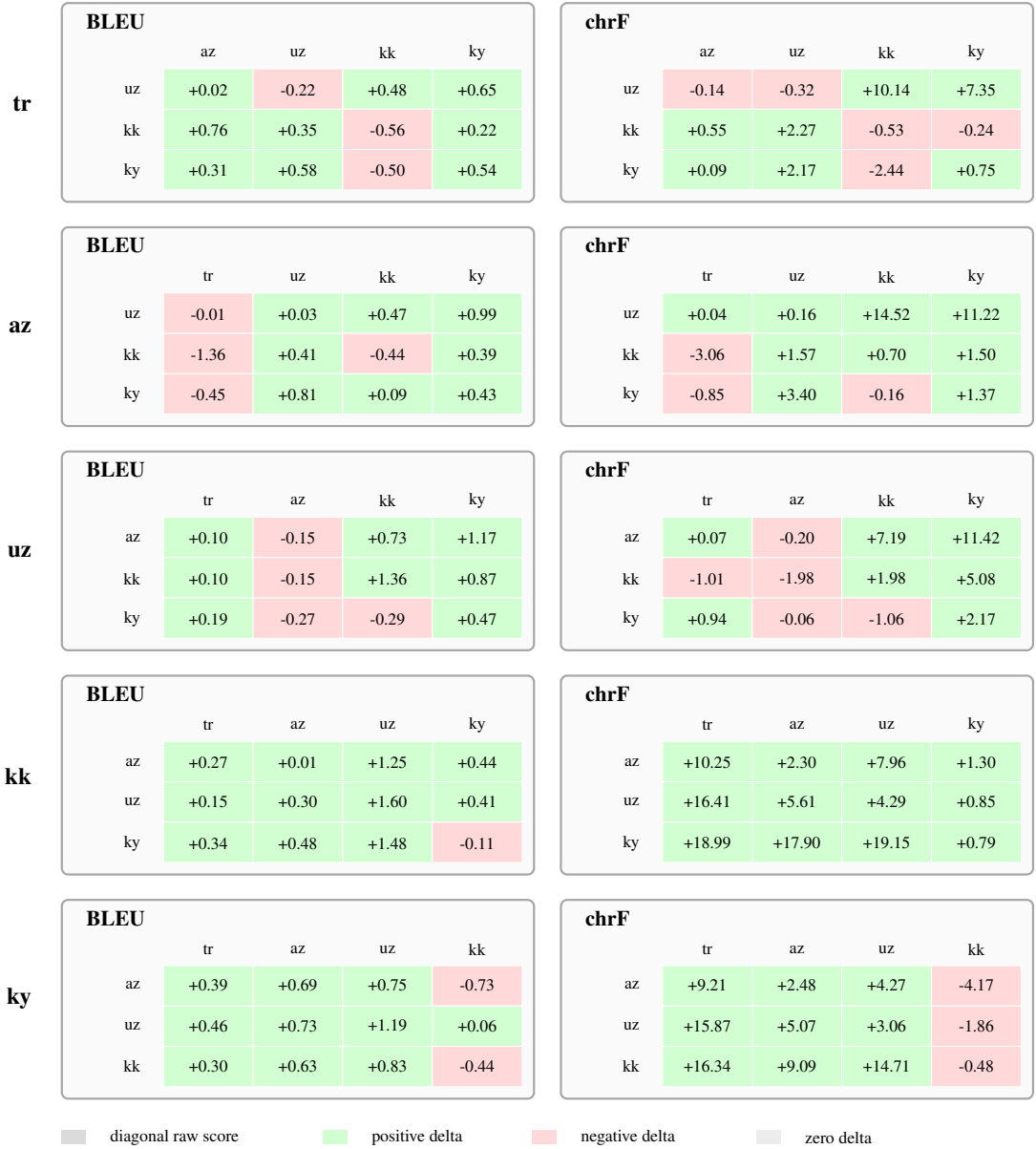

\centering
\begin{adjustbox}{width=\textwidth,max totalheight=0.96\textheight,center}

\end{adjustbox}
\caption{Latinized-minus-original allCPT XWMT raw-score deltas. Rows of panels correspond to fixed translation targets and metric columns correspond to BLEU, chrF, and COMET. Positive cells indicate gains from Latinization, while negative cells indicate decreases relative to the original-script setting.}
\label{fig:app-latin-minus-original-exact}
\end{figure*}

Figure~\ref{fig:app-latin-minus-original-exact} reports the raw-score changes obtained by replacing original-script inputs with Latinized inputs in the allCPT XWMT setting. Positive values indicate that Latinization improves the corresponding fine-tuning/evaluation cell, while negative values indicate a decrease. The figure complements the discussion in Section~\ref{sec:latinization-effect} by showing that Latinization does not act as a uniform improvement mechanism. Instead, its effect is concentrated in script-sensitive directions, especially those involving Kazakh and Kyrgyz, and the magnitude of the change differs substantially across metrics.

\subsubsection{XWMT and Tatoeba test dataset Correlation Details}
\label{app:5-4}

Tables~\ref{tab:app-xwmt-tatoeba-corr-coef}--\ref{tab:app-latin-xwmt-tatoeba-ranking-detail-bleu-chrf} provide the detailed evidence behind the cross-dataset stability analysis. Table~\ref{tab:app-xwmt-tatoeba-corr-coef} reports raw-cell Pearson and Spearman correlations between the XWMT and Tatoeba matrices, while Table~\ref{tab:app-xwmt-tatoeba-ranking-summary} summarizes whether the best donor and complete donor ordering are preserved across datasets. Table~\ref{tab:app-latin-xwmt-tatoeba-ranking-detail-bleu-chrf} gives the corresponding target-level donor-order comparisons for BLEU and chrF. Together, these results show that absolute scores may vary between benchmarks, but the donor preference structure is largely stable for surface metrics, especially BLEU and chrF.

\begin{table*}[!tbp]
\centering
\scriptsize
\setlength{\tabcolsep}{1.5pt}
\renewcommand{\arraystretch}{1.03}
\begin{adjustbox}{max width=\textwidth,center}
\begin{tabular}{lllrrr}\toprule\textbf{Setting} & \textbf{Metric} & \textbf{Target} & \textbf{N cells} & \textbf{Pearson r} & \textbf{Spearman rho} \\
\midrule Original & BLEU & tr & 16.00 & 0.9659 & 0.9324 \\ Original & BLEU & az & 9.00 & \negval{-0.0428} & \negval{-0.3000} \\ Original & BLEU & uz & 9.00 & 0.2949 & 0.2667 \\ Original & BLEU & kk & 9.00 & 0.7480 & 0.7667 \\ Original & BLEU & ky & 9.00 & 0.9542 & 0.8833 \\ Original & BLEU & overall & 52.00 & 0.2877 & 0.3401 \\ Original & chrF & tr & 16.00 & 0.9878 & 0.9882 \\ Original & chrF & az & 16.00 & 0.8445 & 0.9059 \\ Original & chrF & uz & 9.00 & 0.2097 & 0.1500 \\ Original & chrF & kk & 9.00 & 0.0188 & 0.3333 \\ Original & chrF & ky & 9.00 & 0.9340 & 0.8167 \\ Original & chrF & overall & 59.00 & 0.3037 & 0.2776 \\ Original & COMET & tr & 16.00 & 0.8236 & 0.7618 \\ Original & COMET & az & 16.00 & 0.4199 & 0.3853 \\ Original & COMET & uz & 16.00 & 0.4455 & 0.5294 \\ Original & COMET & kk & 9.00 & 0.5487 & 0.5000 \\ Original & COMET & ky & 9.00 & 0.9048 & 0.8787 \\ Original & COMET & overall & 66.00 & 0.2453 & 0.2467 \\ Latinized & BLEU & tr & 16.00 & 0.9515 & 0.8529 \\ Latinized & BLEU & az & 9.00 & \negval{-0.0227} & \negval{-0.0667} \\ Latinized & BLEU & uz & 9.00 & 0.0985 & 0.0333 \\ Latinized & BLEU & kk & 9.00 & 0.8810 & 0.9333 \\ Latinized & BLEU & ky & 9.00 & 0.9706 & 0.8333 \\ Latinized & BLEU & overall & 52.00 & 0.1792 & 0.2706 \\ Latinized & chrF & tr & 16.00 & 0.9776 & 0.9647 \\ Latinized & chrF & az & 16.00 & 0.8847 & 0.9176 \\ Latinized & chrF & uz & 9.00 & \negval{-0.3764} & \negval{-0.2510} \\ Latinized & chrF & kk & 9.00 & 0.3287 & 0.2833 \\ Latinized & chrF & ky & 9.00 & 0.8956 & 0.7833 \\ Latinized & chrF & overall & 59.00 & 0.1598 & 0.1403 \\ Latinized & COMET & tr & 16.00 & 0.9480 & 0.9500 \\ Latinized & COMET & az & 16.00 & 0.5940 & 0.7412 \\ Latinized & COMET & uz & 16.00 & 0.4127 & 0.5706 \\ Latinized & COMET & kk & 9.00 & 0.2515 & 0.4333 \\ Latinized & COMET & ky & 9.00 & \negval{-0.2085} & \negval{-0.0667} \\ Latinized & COMET & overall & 66.00 & 0.2084 & 0.2733 \\
\end{tabular}
\end{adjustbox}
\caption{XWMT--Tatoeba raw-cell correlation coefficients.}
\label{tab:app-xwmt-tatoeba-corr-coef}
\end{table*}

\begin{table*}[!tbp]
\centering
\scriptsize
\setlength{\tabcolsep}{3pt}
\renewcommand{\arraystretch}{1.03}
\begin{adjustbox}{max width=\textwidth,center}
\begin{tabular}{lllcccl}\toprule\textbf{Setting} & \textbf{Metric} & \textbf{Compared columns} & \textbf{Same best donor} & \textbf{Same full order} & \textbf{Same self-transfer best donor} & \textbf{Interpretation} \\
\midrule Original & BLEU & 20.00 & 20/20 & 13/20 & 20/20 & Very stable; suitable for coefficient estimation \\ Original & chrF & 20.00 & 20/20 & 17/20 & 20/20 & Very stable; suitable for coefficient estimation \\ Original & COMET & 20.00 & 9/20 & 6/20 & 5/20 & Unstable; use only as weak/secondary signal \\ Latinized & BLEU & 20.00 & 19/20 & 11/20 & 19/20 & Very stable; suitable for coefficient estimation \\ Latinized & chrF & 20.00 & 20/20 & 14/20 & 20/20 & Very stable; suitable for coefficient estimation \\ Latinized & COMET & 20.00 & 17/20 & 11/20 & 16/20 & Less stable; use only as weak/secondary signal \\
\end{tabular}
\end{adjustbox}
\caption{XWMT--Tatoeba donor-ranking stability summary.}
\label{tab:app-xwmt-tatoeba-ranking-summary}
\end{table*}

\begin{table*}[!tbp]
\centering
\fontsize{4.8}{5.3}\selectfont
\setlength{\tabcolsep}{1.4pt}
\renewcommand{\arraystretch}{0.72}
\begin{adjustbox}{max width=\textwidth,max totalheight=0.88\textheight,center}
\begin{tabular}{llllp{0.255\linewidth}p{0.255\linewidth}p{0.13\linewidth}}
\toprule
\textbf{Setting} & \textbf{Metric} & \textbf{Target} & \textbf{Eval source} & \textbf{XWMT donor order} & \textbf{Tatoeba donor order} & \textbf{Verdict} \\
\midrule
Latinized & BLEU & tr & az & az (13.43) > uz (7.48) > ky (7.05) > kk (6.85) & az (14.19) > uz (8.98) > ky (5.35) > kk (5.27) & same full order \\
Latinized & BLEU & tr & uz & uz (6.29) > ky (2.90) > az (2.84) > kk (2.53) & uz (7.76) > ky (1.50) > az (1.29) > kk (1.16) & same full order \\
Latinized & BLEU & tr & kk & kk (4.06) > ky (1.88) > uz (1.65) > az (1.33) & ky (3.31) > kk (2.57) > uz (1.59) > az (1.15) & different best donor \\
Latinized & BLEU & tr & ky & ky (4.04) > kk (1.96) > uz (1.74) > az (1.44) & ky (2.15) > kk (1.39) > uz (0.71) > az (0.54) & same full order \\
Latinized & BLEU & az & tr & tr (12.86) > ky (4.61) > kk (4.60) > uz (4.36) & tr (12.25) > ky (5.90) > uz (4.46) > kk (4.37) & same best donor \\
Latinized & BLEU & az & uz & uz (9.17) > ky (4.27) > kk (3.94) > tr (3.12) & uz (3.54) > ky (1.49) > tr (1.21) > kk (1.05) & same best donor \\
Latinized & BLEU & az & kk & kk (5.50) > ky (2.58) > uz (1.72) > tr (1.34) & kk (4.89) > ky (2.77) > uz (1.53) > tr (1.45) & same full order \\
Latinized & BLEU & az & ky & ky (5.32) > kk (2.43) > uz (2.28) > tr (1.79) & ky (3.61) > kk (1.99) > uz (1.50) > tr (1.16) & same full order \\
Latinized & BLEU & uz & tr & tr (5.59) > az (2.23) > kk (2.15) > ky (1.92) & tr (4.66) > az (1.93) > ky (0.61) > kk (0.48) & same best donor \\
Latinized & BLEU & uz & az & az (6.99) > tr (3.53) > kk (2.64) > ky (2.42) & az (3.47) > tr (1.68) > ky (0.71) > kk (0.61) & same best donor \\
Latinized & BLEU & uz & kk & kk (6.95) > az (2.08) > tr (1.87) > ky (1.72) & kk (6.62) > tr (1.67) > ky (1.66) > az (1.59) & same best donor \\
Latinized & BLEU & uz & ky & ky (3.96) > kk (2.09) > az (1.91) > tr (1.53) & ky (3.85) > kk (1.87) > az (1.77) > tr (1.49) & same full order \\
Latinized & BLEU & kk & tr & tr (3.94) > az (1.50) > ky (1.08) > uz (0.99) & tr (4.30) > az (3.65) > ky (0.63) > uz (0.51) & same full order \\
Latinized & BLEU & kk & az & az (5.23) > tr (3.18) > uz (1.68) > ky (1.47) & az (4.58) > tr (2.47) > ky (1.82) > uz (1.49) & same best donor \\
Latinized & BLEU & kk & uz & uz (8.90) > az (3.58) > tr (3.16) > ky (2.92) & uz (6.55) > az (3.10) > tr (2.91) > ky (2.36) & same full order \\
Latinized & BLEU & kk & ky & ky (6.12) > az (2.79) > tr (2.67) > uz (2.54) & ky (3.85) > tr (1.41) > uz (1.30) > az (0.95) & same best donor \\
Latinized & BLEU & ky & tr & tr (3.44) > az (1.63) > uz (1.38) > kk (1.28) & tr (1.57) > az (0.95) > uz (0.37) > kk (0.34) & same full order \\
Latinized & BLEU & ky & az & az (4.68) > tr (2.82) > uz (2.15) > kk (1.85) & az (3.18) > tr (2.15) > uz (1.12) > kk (0.97) & same full order \\
Latinized & BLEU & ky & uz & uz (5.85) > az (2.17) > kk (2.15) > tr (2.08) & uz (5.24) > kk (2.41) > tr (2.36) > az (2.06) & same best donor \\
Latinized & BLEU & ky & kk & kk (6.70) > uz (2.66) > az (1.96) > tr (1.91) & kk (3.14) > uz (1.35) > az (1.33) > tr (0.93) & same full order \\
\midrule
Latinized & chrF & tr & az & az (48.64) > uz (40.06) > kk (38.65) > ky (38.26) & az (42.86) > uz (35.10) > kk (32.36) > ky (31.57) & same full order \\
Latinized & chrF & tr & uz & uz (39.45) > az (29.86) > ky (29.68) > kk (29.66) & uz (33.90) > kk (22.40) > ky (22.31) > az (21.48) & same best donor \\
Latinized & chrF & tr & kk & kk (34.70) > ky (27.32) > uz (25.97) > az (24.67) & kk (24.47) > ky (20.40) > uz (18.59) > az (16.97) & same full order \\
Latinized & chrF & tr & ky & ky (33.25) > kk (28.25) > uz (25.37) > az (24.41) & ky (22.50) > kk (21.62) > uz (19.44) > az (18.77) & same full order \\
Latinized & chrF & az & tr & tr (47.74) > kk (32.88) > ky (32.22) > uz (31.87) & tr (41.10) > ky (29.47) > uz (28.82) > kk (28.52) & same best donor \\
Latinized & chrF & az & uz & uz (43.77) > ky (32.83) > kk (32.52) > tr (31.02) & uz (32.44) > ky (24.87) > kk (24.28) > tr (23.43) & same full order \\
Latinized & chrF & az & kk & kk (38.07) > ky (28.58) > uz (25.42) > tr (22.96) & kk (34.35) > ky (27.34) > uz (23.66) > tr (22.26) & same full order \\
Latinized & chrF & az & ky & ky (36.32) > kk (29.26) > uz (24.65) > tr (24.41) & ky (33.68) > kk (27.37) > uz (23.21) > tr (23.09) & same full order \\
Latinized & chrF & uz & tr & tr (38.31) > az (26.77) > kk (23.68) > ky (23.27) & tr (30.43) > az (22.02) > kk (17.47) > ky (17.21) & same full order \\
Latinized & chrF & uz & az & az (41.58) > tr (31.28) > ky (27.50) > kk (27.41) & az (30.77) > tr (23.45) > kk (21.69) > ky (21.08) & same best donor \\
Latinized & chrF & uz & kk & kk (40.31) > az (23.62) > tr (22.05) > ky (21.27) & kk (38.79) > az (23.83) > tr (23.30) > ky (21.74) & same full order \\
Latinized & chrF & uz & ky & ky (34.83) > kk (25.81) > az (24.71) > tr (23.36) & ky (34.64) > kk (26.66) > az (24.93) > tr (23.63) & same full order \\
Latinized & chrF & kk & tr & tr (35.25) > az (24.31) > ky (21.47) > uz (21.16) & tr (28.20) > az (21.08) > ky (17.11) > uz (16.60) & same full order \\
Latinized & chrF & kk & az & az (39.18) > tr (32.02) > uz (25.64) > ky (25.50) & az (35.98) > tr (29.50) > ky (25.36) > uz (24.57) & same best donor \\
Latinized & chrF & kk & uz & uz (46.07) > az (33.56) > tr (31.93) > ky (30.88) & uz (40.46) > az (29.80) > tr (28.33) > ky (28.15) & same full order \\
Latinized & chrF & kk & ky & ky (41.58) > az (32.87) > uz (31.54) > tr (30.92) & ky (29.86) > uz (23.76) > az (23.59) > tr (22.15) & same best donor \\
Latinized & chrF & ky & tr & tr (31.62) > az (23.82) > uz (22.09) > kk (21.58) & tr (22.67) > az (20.17) > uz (18.56) > kk (18.26) & same full order \\
Latinized & chrF & ky & az & az (34.48) > tr (28.53) > uz (25.17) > kk (23.70) & az (34.98) > tr (28.51) > uz (24.86) > kk (23.49) & same full order \\
Latinized & chrF & ky & uz & uz (37.44) > az (26.56) > kk (26.42) > tr (26.33) & uz (38.02) > kk (27.63) > tr (26.62) > az (26.51) & same best donor \\
Latinized & chrF & ky & kk & kk (39.41) > uz (28.91) > az (27.01) > tr (25.15) & kk (27.55) > uz (22.96) > az (21.96) > tr (19.90) & same full order \\
\bottomrule
\end{tabular}
\end{adjustbox}
\caption{Detailed XWMT--Tatoeba donor-order comparisons for the Latinized mT5 setting using BLEU and chrF.}
\label{tab:app-latin-xwmt-tatoeba-ranking-detail-bleu-chrf}
\end{table*}

\subsubsection{Nominal Deltas for noCPT-allCPT Raw Scores}
\label{app:5-6}

\begin{figure*}[!t]
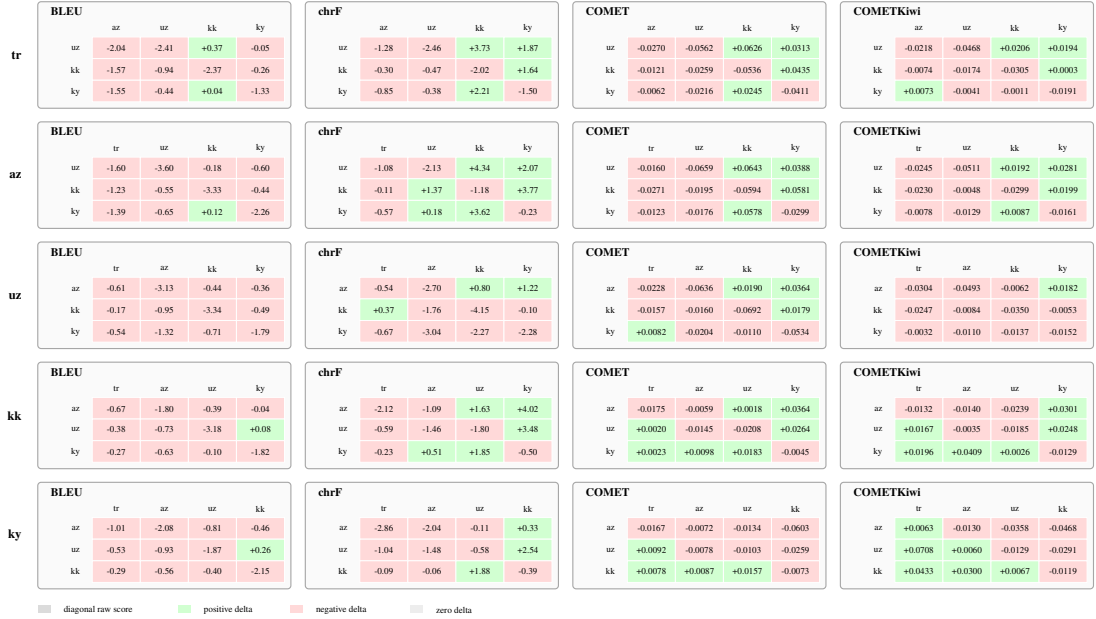

\centering
\begin{adjustbox}{width=\textwidth,max totalheight=0.96\textheight,center}

\end{adjustbox}
\caption{allCPT-minus-noCPT raw-score deltas on Latinized XWMT. Rows of panels correspond to fixed translation targets and metric columns correspond to BLEU, chrF, COMET, and COMETKiwi. Positive cells indicate gains from allCPT, while negative cells indicate stronger noCPT performance.}
\label{fig:app-allcpt-minus-nocpt-exact}
\end{figure*}

Figure~\ref{fig:app-allcpt-minus-nocpt-exact} compares the Latinized XWMT raw-score matrices obtained with allCPT against the corresponding noCPT matrices. Positive values indicate that continual pretraining on all five Turkic languages improves a cell, while negative values indicate that the noCPT model performs better. The comparison shows that allCPT is not uniformly beneficial. The largest decreases often appear on diagonal supervised settings, including BLEU drops of $-2.93$ for az$\rightarrow$tr, $-3.60$ for uz$\rightarrow$az, $-3.34$ for kk$\rightarrow$uz, $-3.18$ for uz$\rightarrow$kk, and $-2.15$ for kk$\rightarrow$ky. At the same time, several kk/ky-related cells improve, such as the target=az chrF gains for kk and ky ($+4.34$ and $+2.07$). This pattern is consistent with the ``curse of multilinguality'', where multilingual training under fixed model capacity can hurt some high-resource or already well-represented directions while benefiting lower-resource ones \citep{chang-etal-2024-multilinguality}. Because mT5-small has limited capacity, shared Turkic adaptation may help underrepresented patterns while degrading stronger supervised representations, especially on diagonal cells.

\subsection{Qwen3 0.6B Recovery Tables}
\label{app:5-9}

\begin{figure*}[!t]
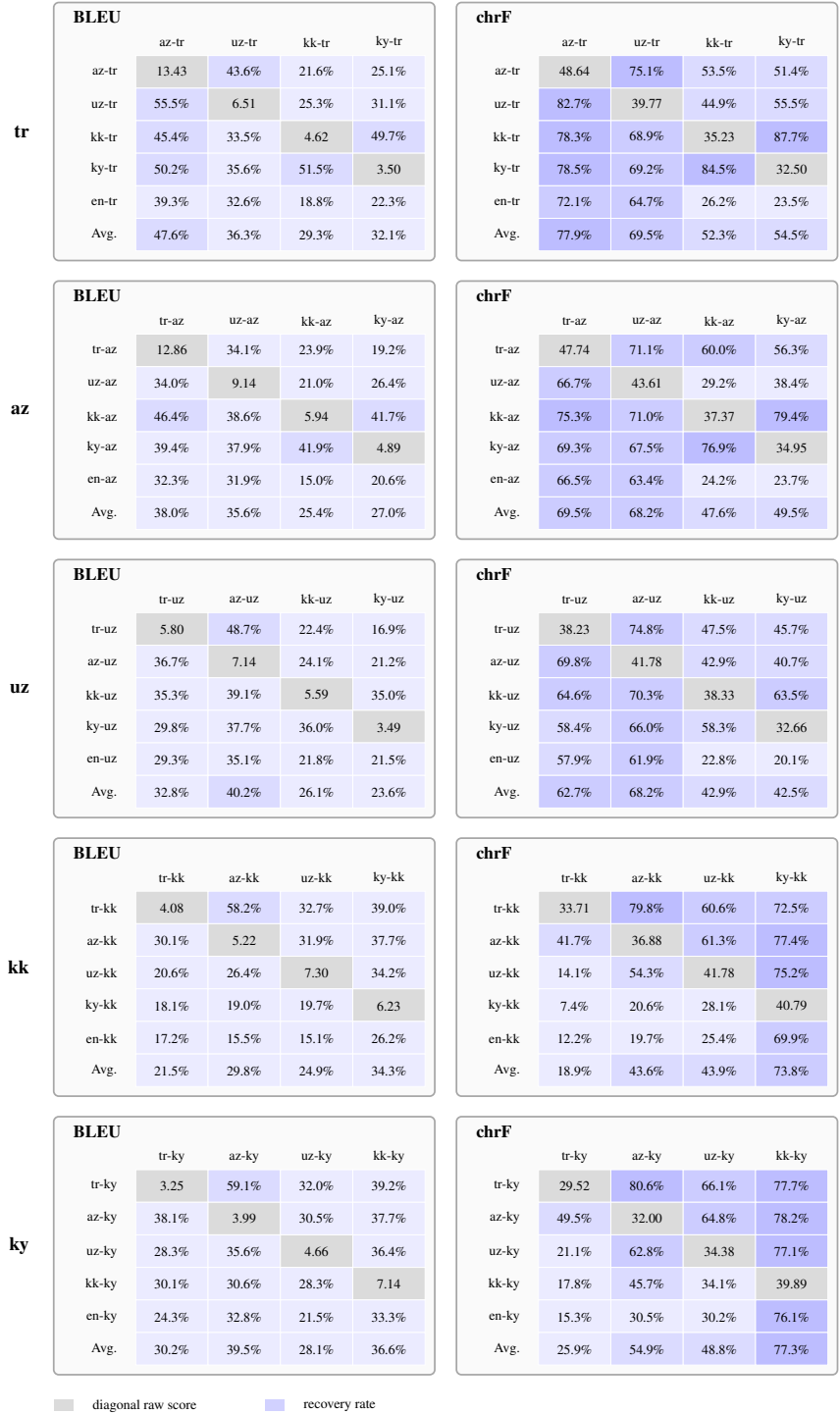

\centering
\begin{adjustbox}{width=\textwidth,max totalheight=0.94\textheight,center}

\end{adjustbox}
\caption{Qwen3 0.6B recovery matrices. Rows of panels correspond to fixed translation targets and metric columns correspond to BLEU and chrF. Diagonal cells report raw same-source scores, off-diagonal cells report recovery rates, and the Avg. row reports average recovery for each fine-tuning source.}
\label{fig:app-qwen-recovery-exact}
\end{figure*}

Figure~\ref{fig:app-qwen-recovery-exact} reports Qwen3 0.6B recovery matrices in the same fixed-target format used for the mT5 experiments. Diagonal cells give the raw same-source score, and off-diagonal cells give the recovery rate for cross-source transfer. These matrices provide architecture-level supporting evidence for the main results: although Qwen3 and mT5 differ in model family and training formulation, the Qwen3 matrices still exhibit structured intra-family transfer rather than random behavior. The figure therefore supports that the observed transfer structure is not merely an artifact of the mT5 encoder--decoder architecture.

\end{document}